\documentclass{bmvc2k}

\usepackage{graphicx}

\usepackage{amsmath}
\usepackage{amssymb}

\graphicspath{{images/}}
\usepackage{svg}
\usepackage{soul}
\usepackage{lipsum}

\title{JEAN: Joint Expression and Audio-guided NeRF-based Talking Face Generation}

\newcommand{\MethodName}{JEAN\xspace}

\addauthor{Sai Tanmay Reddy Chakkera}{schakkera@cs.stonybrook.edu}{1}
\addauthor{Aggelina Chatziagapi}{aggelina@cs.stonybrook.edu}{1}
\addauthor{Dimitris Samaras}{samaras@cs.stonybrook.edu}{1}

\addinstitution{
 Stony Brook University\\
 NY, USA
}

\runninghead{S. Chakkera, A. Chatziagapi, D. Samaras}{JEAN}

\def\eg{\emph{e.g}\bmvaOneDot}
\def\ie{\emph{i.e}\bmvaOneDot}

\begin{document}

\maketitle

\begin{abstract}
  % \documentclass[../main.tex]{subfiles}
% \graphicspath{{\subfix{../images/}}}
% \begin{document}

% We propose a novel method that simultaneously combines lip-syncing and facial expression transfer for talking face generation.
We introduce a novel method for joint expression and audio-guided talking face generation.
Recent approaches 
% that control facial expressions and simultaneously synchronize the lip motion to a target audio, 
either struggle to preserve the speaker identity or fail to produce faithful facial expressions. 
% Other methods have used NeRFs for the task of lip-syncing, and expression transfer separately. 
To address these challenges, we propose a NeRF-based network. Since we train our network on monocular videos without any ground truth, it is essential to learn disentangled representations for audio and expression.
% conditioned on well-designed disentangled representations for audio and expression.
% Our network is trained on monocular talking face videos without any ground truth, and thus requires conditioning on disentangled representations for audio and expression.
We first learn audio features in a self-supervised manner, given utterances from multiple subjects.
% , training an autoencoder on  multiple subjects.
By incorporating a contrastive learning technique, we ensure that the learned audio features are aligned to the lip motion and disentangled from the muscle motion of the rest of the face. We then devise a transformer-based architecture that learns expression features, capturing long-range facial expressions and disentangling them from the speech-specific mouth movements. 
% The corresponding learned representations for audio and expression condition our proposed dynamic NeRF.
% In this work, we present a dynamic NeRF that performs expression-guided lip-synchronization. We devise a self-supervised expression disentanglement training regime that disentangles expression content from the mouth motion content in expression features. Further, we propose a contrastive audio encoder training regime that aligns the audio feature space to the mouth motion feature space. This leads to better lip-sync accuracy. Combining the audio encoder pretraining with a disentangled expression feature space gives us a talking face generator that is faithful to input expressions. 
% an expression controllable neural talking face generator. 
% In brief, the contributions of our work are: (a) a novel NeRF-based method that performs expression-guided lip synchronization, (b) an expression disentanglement module that disentangles expression-specific motion from speech-specific motion, and (c) a self-supervised audio encoder method to align the audio feature space to the lip motion feature space that improves lip-synchronization on unseen audio. 
Through quantitative and qualitative evaluation, we demonstrate that our method can synthesize high-fidelity talking face videos, achieving state-of-the-art facial expression transfer along with lip synchronization to unseen audio. Project Page: \href{https://starc52.github.io/publications/2024-07-19-JEAN}{https://starc52.github.io/publications/JEAN}

% \end{document}
\end{abstract}

%-------------------------------------------------------------------------
\section{Introduction}
\label{sec:intro}
% \documentclass[../main.tex]{subfiles}
% \graphicspath{{\subfix{../images/}}}
% \begin{document}

Talking face generation has increasingly drawn attention due to its wide-ranging applications such as visual dubbing, video content creation and video conferencing. There are two main requirements in synthesizing a photorealistic talking face: (a) accurate lip synchronization to the spoken utterance, and (b) faithful facial expressions to convey a message with the intended affect. In human interaction, facial expressions deliver essential cues while talking~\cite{facs, whatfacereveals}. For example, the same sentence spoken with an angry or happy emotion can have a different meaning.
% To produce lifelike emotions, faithful expression generation is necessary. 
Prior work has mostly focused on audio-only~\cite{wav2lip, makeittalk, neuralvoicepup, speech2vid, sda, pc-avs} or expression-only~\cite{pumarola2018ganimation, defgan, deepvidportraits, Doukas2020Head2HeadDF, Koujan2020Head2HeadVN, Athar2020FaceDet3DFE, det3d} guidance for face synthesis.
% has also addressed expression driven face generation. Emotional expressions are necessary to deliver essential cues while talking~\cite{whatfacereveals}. Hence, 
A few methods~\cite{jieamm, wang2022pdfgc, jang2023thats, evp} have tried to address the problem of simultaneous control of facial expressions and lips.
% for talking face generation.
However, they either struggle to preserve the speaker identity or fail to produce faithful expressions.
% In contrast, other methods have used
Recently, neural radiance fields (NeRFs)~\cite{mildenhall2020nerf} have demonstrated photorealistic 3D modeling, preserving identity-specific information and faithfully reconstructing expressions~\cite{nerface}. However, NeRF-based methods have only addressed the problems of lip-syncing~\cite{guo2021adnerf, ssp-nerf, li2023ernerf, lipnerf, ye2023geneface}
% , tang2022radnerf, yao2022dfa, ye2023geneface++, aenerf, sdnerf}
or expression transfer \cite{nerface, flame-nerf, park2021hypernerf, rig-nerf} separately.
% , but not together. 
  
In this work, we present \MethodName, a novel \textbf{J}oint \textbf{E}xpression and \textbf{A}udio-guided \textbf{N}eRF for talking face generation.
% expression-guided lip-synchronization. 
Our network is trained on monocular talking face videos without any ground truth. In these videos, the expression-related facial movements are strongly entangled with the speech-specific mouth movements. 
Controlling facial expressions and speech-specific lip motion separately
% in a disentangled manner
requires learning \emph{disentangled} representations for expression and audio correspondingly.
% generating the dynamic facial movements, which are implicitly but strongly coupled with mouth movements.
% This implies that there is a conflict between depicting a particular facial expression and articulating lip-motion associated to a particular audio.
To address this, we introduce a self-supervised approach to disentangle facial expressions from lip motion. 
We observe that mouth motion related to speech and face motion related to expressions in talking faces differ from each other temporally and spatially. Speech-related motion has higher temporal frequency and is spatially localized to the mouth region, while expression-related face motion has a lower temporal frequency and may occur over the entire face region. Moreover, for the same utterance spoken with different expressions, speech-related motion remains consistent. We leverage these observations to disentangle speech-related motion and expression-related motion.

\begin{figure}[t]
    \centering
    \includegraphics[width=\textwidth]{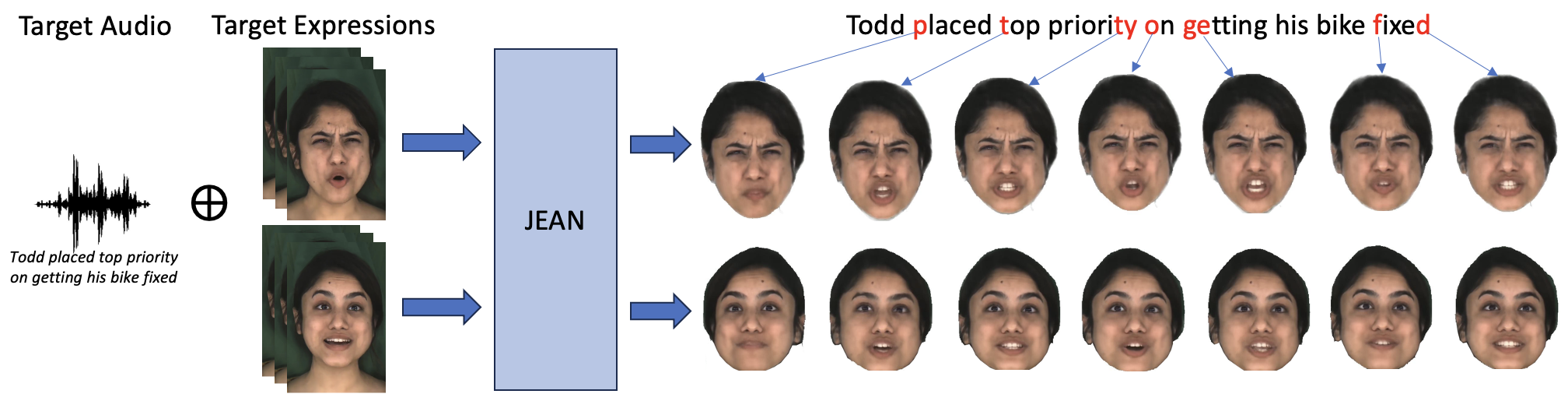}
    \caption{We introduce \MethodName, a novel NeRF-based method that simultaneously combines lip-syncing to a target audio with facial expression transfer to generate talking faces.}
    \label{fig:teaser-caption}
    \vspace{-15pt}
\end{figure}

We first learn a powerful audio representation in a self-supervised manner by disentangling the lip motion from the motion of the rest of the face in the feature space of an autoencoder. In general, achieving accurate lip synchronization on unseen audio in NeRFs is hard, as they tend to overfit on the training data~\cite{guo2021adnerf}. Recently, contrastive learning has shown promise in synchronization in audio-visual tasks~\cite{davs, wu2023speech2lip, Wang_2023_CVPR, selftalk}. This prompted us to introduce a contrastive learning strategy, in order to align the learned audio features with the lip motion. 
% Specifically, we introduce a method to disentangle lip motion from eye motion in the feature space. 
% Then, we align these lip-motion features to audio features.
Next, we introduce a transformer-based architecture that learns expression features, capturing long-range facial expressions and disentangling them from speech-specific lip motion.
Finally, we train a dynamic NeRF, conditioned on the learned representations for expression and audio. \MethodName can synthesize high-fidelity talking face videos, faithfully following both the input facial expressions and speech signal for a given identity. 
% our disentangled expression features and aligned audio features that generates faithful facial expressions without hurting lip-sync.

In brief, the contributions of our work are as follows:
\vspace{-8pt}
\begin{itemize}
    % \item We propose a novel NeRF-based method that performs simultaneous expression control and lip synchronization.
    \item We introduce a self-supervised method to extract audio features aligned to lip motion features,
    % of an autoencoder,
    achieving accurate lip synchronization on unseen audio.
    \vspace{-8pt} 
    \item We propose a transformer-based module to learn expression features, disentangled from speech-specific mouth motion.
    % disentanglement module that disentangles expression-specific motion from speech-specific motion.
    \vspace{-8pt}
    \item Conditioning on the disentangled representations for expression and audio, we propose a novel NeRF-based method for simultaneous expression control and lip synchronization, outperforming the current state-of-the-art.
    % performs simultaneous expression control and lip synchronization
    % We demonstrate our method's superiority over the current state-of-the-art by achieving better expression control and lip-sync accuracy than the state-of-the-art.
\end{itemize}

% \end{document}

\section{Related Work}
\label{sec:related_work}
% \documentclass[../main.tex]{subfiles}
% \graphicspath{{\subfix{../images/}}}
% \begin{document}

\textbf{Audio-driven Talking Face Generation.}
Audio-driven talking face generation aims to generate portrait images with synchronized lip motion to a given speech. Early attempts in talking face generation with lip synchronization~\cite{Sako2000HMMbasedTS, 1407897, 10.1145/2783258.2783356} use probabilistic models to map speech phonemes
% in audio
to particular mouth shapes, requiring accurate annotation. More recent methods~\cite{syn-obama, speech2vid,makeittalk, neuralvoicepup,sda,atvg,davs,pc-avs,wav2lip} learn neural networks, such as GANs, using a large amount of video data, containing multiple identities, in order to learn a robust audio-lip space.
% in video (visemes). 
% using HMMs~\cite{Sako2000HMMbasedTS}, decision trees~\cite{1407897} and LSTMs~\cite{10.1145/2783258.2783356}. 
% However, these methods require very accurate labelling of phonemes (at millisecond level) using automatic speech recognition, which is often error prone. More recent methods avoid such explicit mapping between audio (phonemes) and video (visemes)~\cite{syn-obama, speech2vid, makeittalk, thies2020neural}. 
% Synthesizing Obama \cite{syn-obama} synthesizes talking face videos of Obama using $\approx 17$ hours of video data. Speech2Vid \cite{DBLP:journals/corr/ChungJZ17} uses an encoder-decoder architecture, where each input face image is conditioned on the corresponding speech segment. MakeItTalk~\cite{makeittalk} and Neural Voice Puppetry \cite{thies2020neural} use intermediate 3D representations to learn a mapping between audio and video. 
% GAN-based approaches~\cite{sda,atvg,davs,pc-avs,wav2lip}
% , such as SDA \cite{sda}, ATVG \cite{atvg}, DAVS \cite{davs}, PC-AVS~\cite{pc-avs}, Wav2Lip \cite{wav2lip},
% usually require a large amount of data containing multiple identities, in order to learn a robust audio-lip space.
% use 2D intermediate representations.
Our method, on the other hand, is NeRF-based, which enables us to better capture the 3D geometry and appearance of a talking face, and achieve higher output visual quality 
% in the output,
with just a few minutes of monocular video data. 

% GAN-based approaches usually require a large amount of data containing multiple identities, in order to learn a robust audio-lip space. SDA \cite{sda}, ATVG \cite{chen2019hierarchical}, DAVS \cite{davs}, PC-AVS~\cite{pc-avs}, Wav2Lip \cite{wav2lip} are such GAN-based approaches that use intermediate 2D representations. While these methods give competitive lip-synchronization accuracy, they are generally limited in resolution, and are 2D in nature leading to 3D-inconsistency and limits in pose variability. 

\noindent
\textbf{NeRFs for Human Faces.}
Implicit neural representations for modeling 3D scenes have recently gained a lot of attention. In particular, neural radiance fields (NeRFs)~\cite{mildenhall2020nerf} have shown photorealistic novel view synthesis of complex static~\cite{barron2021mip, barron2022mip} and dynamic~\cite{Li_2022_CVPR, li2020neural,pumarola2021d,xian2021space} scenes.
% both static and dynamic.
% NeRF-based methods
They represent a scene using an MLP, where each 3D point is associated with a radiance and density.
Various recent works~\cite{nerface, park2021nerfies, park2021hypernerf, raj2020pva, sun2021nelf} use NeRFs to model the 3D face geometry.
% More recently, NeRFs have made it possible to model the 3D face geometry more accurately~\cite{nerface, park2021nerfies, park2021hypernerf, raj2020pva, sun2021nelf}, leading to more photorealistic results. 
% Recent NeRF based approaches are in contrast, in that, these represent a scene by mapping each point in 3D space to a particular colour and density giving them the ability to model 3D face geometry more accurately. 
AD-NeRF~\cite{guo2021adnerf} learns a dynamic NeRF conditioned on speech, encoded as DeepSpeech features~\cite{hannun2014deep,deepspeech}.
% , but performs poorly in case of novel audio.
% . While AD-NeRF performs generally well in generating lip-synced talking heads of high visual quality, it performs poorly in case of unseen audio. 
% (any audio other than the one it trains on), producing unnatural lip-synchronization. 
Follow-up methods~\cite{ssp-nerf, yao2022dfa, ye2023geneface, ye2023geneface++, tang2022radnerf, li2023ernerf, aenerf, sdnerf} enhance the lip synchronization in case of novel audio.
% in such cases. 
% SSP-NeRF~\cite{ssp-nerf} learns a semantic-aware dynamic ray sampling to better learn face regions corresponding to speech. Yao \etal~\cite{yao2022dfa}, GeneFace~\cite{ye2023geneface}, GeneFace++~\cite{ye2023geneface++}, RADNeRF~\cite{tang2022radnerf},~\cite{li2023ernerf} propose improvements on lip synchronization in the unseen audio case. AE-NeRF~\cite{aenerf} addresses the task of few-shot talking face generation. SD-NeRF~\cite{sdnerf} incorporates facial movements like blinks and brow movements. 
 % ~\cite{rig-nerf, flame-nerf, park2021hypernerf}
NeRFace~\cite{nerface} conditions the network on 3DMM expression parameters. In contrast to the aforementioned approaches, our proposed NeRF allows for simultaneous control over facial expressions and lip motion to unseen audio.

\noindent
\textbf{Representation Learning for Human Faces.}
The task of representation learning for faces has been widely explored in unsupervised~\cite{chen2018isolating,chen2016infogan,ding2020guided,higgins2017betavae,kim2018disentangling,nguyen2019hologan,wei2021orthogonal} and self-, semi-, or weakly-supervised~\cite{deng2020disentangled,Ghosh2020GIFGI,pumarola2018ganimation,Ren2021PIRendererCP} techniques.
% Unsupervised methods lead to controllable face features that are not semantically meaningful~\cite{locatello2019challenging}. To add semantic meaning to learned representations supervision is necessary.
In the talking face setting, where supervision is scarce and hard to find, self-supervision has been widely explored in the literature. Some methods have used self-supervision to improve the lip synchronization~\cite{Wang_2023_CVPR, selftalk, yao2022dfa} in talking head generation tasks. Other methods have used self-supervised learning~\cite{invarmotion, Gao_2023_CVPR, wang2022pdfgc} to disentangle pose, expression, eye motion, etc., of a talking face to enable individual control. Other methods have used self-supervision to learn proxies, such as depth~\cite{hong2023dagan, hong2022depth}, latent features~\cite{latentavatar} or keypoints~\cite{jieamm} that improve generated talking faces. Our method disentangles expression from lip motion and aligns audio features to lip motion using self-supervision.

\noindent
\textbf{Expression and Audio-driven Talking Face Generation.}
 % (parameters such as language of source, gender, etc affect the ``natural'' look of the rendered lip synchronization). This poor performance can be attributed to the overfitting of the NeRF on a training video entangled representation of speech. Hence, learning better a audio representation for lip-sync which is aligned to the mouth positions is necessary. 
Expression and audio-driven talking face generation aims to produce portrait images that follow expressions from an expression source and lips synchronized to an audio source. Prior work can be broadly classified into warping-based methods~\cite{jieamm, emmn} or synthesis-based methods~\cite{wang2022pdfgc, stochlattfg, jang2023thats, evp, goyal2023emotionally, Sinha2022EmotionControllableGT, peng2023synctalk}. Warping-based techniques estimate warping flows between source and target images,
% to transform the target,
whereas synthesis-based methods generate images based on intermediate representations. Warping-based techniques, like EAMM~\cite{jieamm}, frequently produce 3D inconsistencies, since they consider only the 2D space. Synthesis-based methods, like PD-FGC~\cite{wang2022pdfgc}, lead to the semantic leakage problem~\cite{xia2023gmtalker}, where the output erroneously contains semantic elements of the input or training dataset. Some methods~\cite{stochlattfg, goyal2023emotionally, Sinha2022EmotionControllableGT}, such as EAT~\cite{EAT_gen}, learn a categorical emotional space, often based on one-hot encodings.
% to construct emotional expressions. 
% Other methods such as EVP~\cite{evp} and EMMN~\cite{emmn} disentangle emotion information from the audio signal to drive emotional expressions in generated videos. 
% Unlike aforementioned generative methods, SyncTalk~\cite{peng2023synctalk} incorporates expression control in a NeRF based formulation to make the audio dubbing more realistic by enabling control of eyebrows, eyes and face. However, their formulation does not allow for explicit control of expressions from unseen sources or handling different emotions. 
In contrast, our method learns a NeRF controllable by both audio and expression from independent sources, giving 3D accurate and identity-preserved outputs with faithful expressions. 

% learn key-points in a self-supervised manner that can be detected and warped on input frames to induce the expression desired while maintaining the lip-synchronization. These were possible due to the large-scale data available at the disposal in these methods. Incorporating such large-scale datasets in NeRFs is hard.

% \subsection{Emotion disentanglement}
% Emotion disentanglement for faces is a longstanding task and has been widely explored in the literature. Previous works have mostly focused on audio emotion disentanglement  

% \end{document}
% \vspace{-18pt}

\section{Method}
\label{sec:method}
% \documentclass[../main.tex]{subfiles}
% \graphicspath{{\subfix{../images/}}}
% \begin{document}

% \subsection{Overview}

\begin{figure*}
  \includegraphics[width=\textwidth]{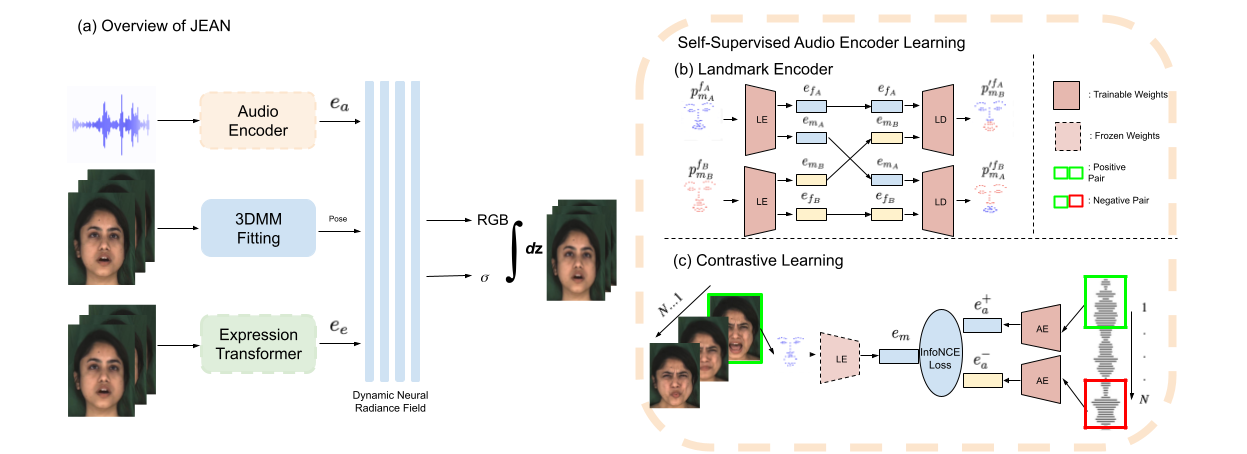}
  \caption{(a) illustrates an overview of \MethodName, a novel method for joint expression and audio-guided NeRF-based 
  % that learns an expression-guided dynamic NeRF for 
  talking face generation.
  % of our proposed technique.
  (b) and (c) illustrate our proposed self-supervised learning of our audio representation. 
  % We propose a self-supervised learning method to align audio features to lip motion.
  Specifically, (b) demonstrates the self-supervised learning of our landmark autoencoder that disentangles lip motion from the motion of the rest of the face. Then, in (c), our audio encoder $AE$ is trained using a contrastive learning regime, in order to align audio features to lip motion.}
  % features.}
  % using a contrastive learning regime guided by InfoNCE loss. }
  \label{fig:model}
  \vspace{-10pt}
\end{figure*}

We present \MethodName, a novel method for joint expression and audio-guided NeRF-based 
% learns an expression-guided dynamic NeRF for
talking face generation. 
Fig.~\ref{fig:model}(a) illustrates an overview of our proposed approach. Given monocular RGB videos of an identity, we learn a NeRF that represents the identity's 4D face geometry and appearance in various expressions and lip positions. We assume three inputs, namely an audio source, the identity's head pose, and an expression source. During training, these inputs come from the same identity. During inference, we can use audio, pose, and expression sources from different videos. Our proposed pipeline consists of three main components:
(1) We first learn an audio encoder in a self-supervised manner to align audio features to lip motion features (see Sec.~\ref{sec:method_audio}).
(2) We learn an expression transformer to disentangle expression features from lip motion (see Sec.~\ref{sec:method_exp}).
(3) Finally, we learn a dynamic NeRF conditioned on our learned representations for both audio and expression (see Sec.~\ref{sec:method_nerf}).

\subsection{Self-Supervised Audio Encoder}\label{sec:method_audio}

% In order to improve lip-sync accuracy, we propose a self-supervised learning method to align audio features to lip motion.
% , while keeping it disentangled from mouth positions.
% Such disentangled representations for lip-sync can be learned using visual cues as suggested in DFA-NeRF \cite{yao2022dfa}. Inspired by their formulation, we learn an audio encoder to align to disentangled lip-motion features. 

In order to learn a powerful audio representation and achieve a high lip-sync accuracy, we propose a self-supervised contrastive learning method that aligns audio features to lip motion features. Inspired by Yao et al.~\cite{yao2022dfa}, we first extract lip motion features through a landmark autoencoder.
% , which disentangles mouth from eye-nose movements. 
% (everything excluding the mouth).
Next, we train our audio encoder using a contrastive learning strategy.
% use the learned lip-motion features in a contrastive learning strategy, in order to train our audio encoder.

\noindent
\textbf{Landmark Autoencoder.}
We propose a landmark autoencoder that learns to disentangle mouth and eye-nose movements based on 2D landmarks, as illustrated in Fig.~\ref{fig:model}(b). For a frame $A$ of an identity, we extract face landmarks $\textbf{p}^{f_A}_{m_A}$,
% where $\textbf{p}^{f_A}_{m_A}$ indicates all landmark points
with superscript $f_A$ indicating that the eye-nose landmarks are from frame $A$ and subscript $m_A$ indicating that mouth landmarks are from frame $A$. Similarly, for a frame $B$, we extract face landmarks $\textbf{p}^{f_B}_{m_B}$. A landmark encoder (\textit{LE}) embeds the input landmarks of each frame into a eye-nose embedding $\textbf{e}_{f}$ and a mouth embedding $\textbf{e}_{m}$, \ie $\textbf{e}_{f_A}, \textbf{e}_{m_A}=\text{\textit{LE}}(\textbf{p}^{f_A}_{m_A})$ and $\textbf{e}_{f_B}, \textbf{e}_{m_B}=\text{\textit{LE}}(\textbf{p}^{f_B}_{m_B})$. 
% The landmarks of two frames $A$ and $B$ are input to the landmark encoder simultaneously.
The mouth embeddings $\textbf{e}_{m_A}$ and $\textbf{e}_{m_B}$ of the two frames $A$ and $B$ correspondingly are swapped with a probability $\epsilon$, and passed to a landmark decoder (\textit{LD}) to predict the corresponding landmarks  $\textbf{p}^{\prime f_A}_{m_B}=\text{\textit{LD}}(\textbf{e}_{f_A}, \textbf{e}_{m_B})$ and $\textbf{p}^{\prime f_B}_{m_A}=\text{\textit{LD}}(\textbf{e}_{f_B}, \textbf{e}_{m_A})$. We get the ground truth $\textbf{p}^{f_A}_{m_B}$ by replacing the mouth landmarks of the frame $A$ with the corresponding mouth landmarks of the frame $B$. The autoencoder is trained using an L1 reconstruction loss:
% is computed using an L1 loss. Specifically, the reconstruction loss is 
\begin{equation}
    \mathcal{L}_{{rec}_{lmd}} = \mathbf{E}\left[||\textbf{p}^{\prime f_A}_{m_B}-\textbf{p}^{f_A}_{m_B}||_1 + ||\textbf{p}^{\prime f_B}_{m_A}-\textbf{p}^{f_B}_{m_A}||_1\right].
\end{equation}
Using this training regime, the landmark encoder \textit{LE} learns to represent the lip movements in its latent space, disentangling them from any other face motion.
In our implementation, we extracted 68 face landmarks for each video frame. We discarded the first 17 landmarks that correspond to the face contour, in order to pay attention to the eye-nose and mouth movements. The probability $\epsilon$ is set to $0.8$. The frames $A$ and $B$ are randomly sampled from the same video of an identity (see also suppl.).
% to keep the feature disentanglement identity and video emotion agnostic. Rephrase this: 
% We noticed that sampling from different videos leads to the network learning features that may contain identity and emotion information (see also suppl.). So during training, the frames $A$ and $B$ are randomly sampled from the same video of an identity.

%% Copying this to supplementary
% \paragraph{Implementation details}
% We used 68 point landmarks of which 17 landmarks corresponding to the jaw and edge of face were discarded for training the landmark autoencoder and these were input to the autoencoder as arrays. $\epsilon$ was set to $0.8$. For swapping, the last 20 points in the landmarks representing the lips were used. For this task, we use the complete available set of frontal view MEAD dataset videos, \textit{i.e.} ``part0''. $80\%$ of the samples were kept for training and rest for validation. During training, the frames were are randomly sampled from the same video of an identity to keep the feature disentanglement identity and video emotion agnostic. The landmark encoder and decoder consisted of 4 MLPs each with 256, 256, 128 as the hidden layer sizes. The output features $e_f$ and $e_m$ were 64 dimensions each. The model was optimized using Adam with default parameters and $10^{-5}$ as weight decay value. 

% \subsubsection{Contrastive Learning}

\noindent
\textbf{Contrastive Learning.}
In order to learn audio embeddings aligned to the extracted mouth embeddings $\textbf{e}_m$, we propose a constrastive training strategy, as illustrated in Fig.~\ref{fig:model}(c).
% Based on \cite{yao2022dfa} 
% we use mouth embeddings directly to synchronize auditory utterances. 
We learn a CNN-based audio encoder \textit{AE} that takes DeepSpeech~\cite{hannun2014deep} features $\textbf{a}$ as input and outputs audio embeddings $\textbf{e}_a$, \ie $\textbf{e}_a = \text{\textit{AE}}(\textbf{a})$.
% where $a$ represents the input audio deepspeech features.
% We adopt a contrastive learning strategy to synchronize the mouth embeddings $\textbf{e}_m$ with the audio embeddings $\textbf{e}_a$. Specifically,
For a mouth embedding $\textbf{e}_m$, we set the corresponding audio feature $\textbf{e}^{+}_a$ as the positive key and a randomly picked audio feature $\textbf{e}^{-}_a$ as the negative key. We train our audio encoder, using an InfoNCE~\cite{oord2019representation} loss, 
% for contrastive learning
to ensure that the distance between the positive pair $(\textbf{e}^{+}_a,\textbf{e}_m)$ is smaller 
than the negative one $(\textbf{e}^{-}_a, \textbf{e}_m)$:
% than the one of the negative pair $(\textbf{e}^{-}_a, \textbf{e}_m)$:
% . The InfoNCE loss is defined as:
\begin{equation}
    \mathcal{L}_{\text{InfoNCE}} = -\underset{x \in \mathcal{X}}{\mathbf{E}} \left[ \log{\frac{\exp(d(\textbf{e}^{+}_{a_x}, \textbf{e}_{m_x}))}{\exp(d(\textbf{e}^{+}_{a_x}, \textbf{e}_{m_x}))+\exp(d(\textbf{e}^{-}_{a_x}, \textbf{e}_{m_x}))}}\right],
\end{equation}
where $\mathcal{X}$ is the set of all $(\textbf{e}^{+}_{a}, \textbf{e}_m, \textbf{e}^{-}_{a})$ tuples and $d(\textbf{x}, \textbf{y}) = \frac{\langle \textbf{x}, \textbf{y}\rangle}{\tau ||\textbf{x}||_2||\textbf{y}||_2}$ is the temperature-adjusted cosine distance. During training, the negative audio samples are randomly selected from the same identity but different video. The temperature $\tau$ is set to $0.1$. We use 64-dimensional features for $\textbf{e}_{a}$, $\textbf{e}_{f}$, and $\textbf{e}_{m}$. See suppl.~for more details.

\subsection{Expression Transformer}\label{sec:method_exp}

\begin{figure}
  \centering
  \includegraphics[width=0.75\linewidth]{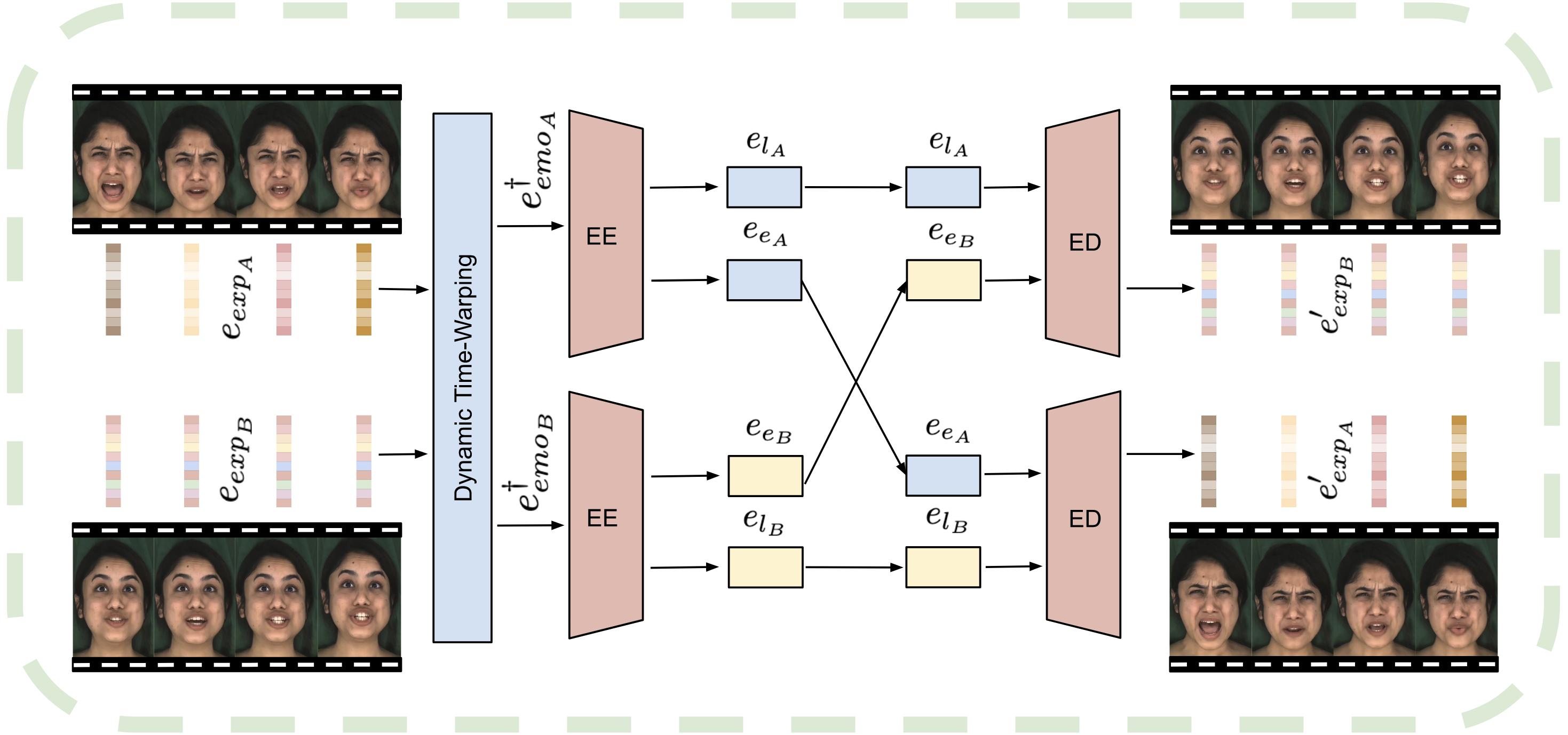}
  \caption{Expression Transformer. We propose an expression transformer encoder that learns to disentangle facial expressions from speech-specific lip motion. We extract emotion features and disentangle them into expression content and speech-specific lip motion content.}
  \label{fig:expr-disent}
  \vspace{-10pt}
\end{figure}

We propose to learn an expression transformer that captures long-range facial expressions, disentangling them from the speech-specific lip motion (see Fig.~\ref{fig:expr-disent}).
First, we extract emotion features $e_{emo}$ per video frame, using a pre-trained network for emotion recognition~\cite{EMOCA:CVPR:2021}.
% We first extract emotion features using a pre-trained ResNet based single image emotion recognition network from ~\cite{EMOCA:CVPR:2021}.
% disentangles expressions from lip motion. We use emotion recognition features from \cite{EMOCA:CVPR:2021}, \cite{DECA:Siggraph2021}
We then learn an expression encoder that disentangles the emotion features into expression content and speech-specific content. The idea is that when a person is speaking, the face movements will have some aspects that are emotion-specific (e.g. cheeks pulled up, flush face, raised eyebrows, etc.) and some speech-specific (e.g. mouth motion for consonant `b'). Given that we have video pairs of a person saying the same utterance in different emotions, it is possible to capture the facial expressions and successfully disentangle them from the speech-specific motion.
% To account for any misalignment of the utterances of different emotions, we incorporate the dynamic time-warping (DTW) algorithm~\cite{1104847} to align the input emotion feature sequences. 
% This method is illustrated in Fig.~\ref{fig:expr-disent}.
More specifically, for an utterance spoken with two different emotions $A$ and $B$, we extract emotion features $\textbf{e}_{{emo}_A[1:m_1]}$ and $\textbf{e}_{{emo}_B[1:m_2]}$. We align these sequences, using the dynamic time warping (DTW) algorithm~\cite{1104847}:
% We take an utterance spoken with two different emotions and extract emotion features $e_{emo}$ from each frame, using the pre-trained ResNet-based emotion recognition network from ~\cite{EMOCA:CVPR:2021}, forming a sequence of emotion features. These features are warped using DTW~\cite{1104847}, in order to be aligned and of the same length. More specifically, the emotion features $\textbf{e}_{{emo}_A[1:m_1]}$ for an utterance in emotion $A$ and the emotion features $\textbf{e}_{{emo}_B[1:m_2]}$ for the corresponding utterance in emotion $B$ are warped as follows:
\begin{equation}
    \textbf{e}^{\dag}_{{emo}_A[1:N]}, \textbf{e}^{\dag}_{{emo}_B[1:N]} = DTW(\textbf{e}_{{emo}_A[1:m_1]}, \textbf{e}_{{emo}_B[1:m_2]}) ,
\end{equation}
where $m_1$ and $m_2$ are the initial lengths and $N$ is the output length of DTW. These are then given as input to the expression transformer encoder ($\text{\textit{EE}}$) in windows of size $\omega$. $\text{\textit{EE}}$ outputs expression features $\textbf{e}_e$ and speech-specific lip motion features $\textbf{e}_l$, \ie $\textbf{e}_l[t:\omega+t-1], \textbf{e}_e[t:\omega+t-1] = \text{\textit{EE}}(\textbf{e}^{\dag}_{{emo}[t:\omega+t-1]})$ where $t \in \mathbb{N}_{1:N}$. The expression features are randomly swapped with a probability $\delta$. The output features $\textbf{e}_e$ and $\textbf{e}_l$ are input to the expression decoder ($\text{\textit{ED}}$). $\text{\textit{ED}}$ follows an auto-regressive architecture to reconstruct the emotion features, \ie $\textbf{e}^{\prime}_{{emo}[t:\omega+t-1]} = \text{\textit{ED}}(\textbf{e}_l[t:\omega+t-1], \textbf{e}_e[t:\omega+t-1])$. The expression transformer is trained using an L1 reconstruction loss:
% is computed between the warped input to the encoder and the reconstructed output of the decoder: 
\begin{equation}
    \mathcal{L}_{{rec}_{emo}} = \mathbf{E}\left[||\textbf{e}^{\prime}_{{emo}_A}-\textbf{e}^{\dag}_{{emo}_A}||_1+||\textbf{e}^{\prime}_{{emo}_B}-\textbf{e}^{\dag}_{{emo}_B}||_1\right].
\end{equation}
During inference, DTW is skipped and the emotion features are directly input to $EE$.
% the expression transformer encoder.
Our expression encoder is identity-specific, capturing each person's unique way of speaking with a particular emotion. In our experiments, we set $\omega = 8$ and $\delta = 0.8$. 
$EE$ and $ED$ have $3$ layers and $8$ attention heads each. The emotion features $e_{emo}$
% The emotion features extracted from the pre-trained emotion recognition network~\cite{EMOCA:CVPR:2021}
are of $2048$ dimension and mapped to $128$ via 2 linear layers.
The output of $EE$ is split in half resulting in $64$-dimensional $\textbf{e}_l$, $\textbf{e}_e$.

\subsection{Dynamic NeRF}\label{sec:method_nerf}

Our learned audio features $\textbf{e}_a$ and expression features $\textbf{e}_e$ are concatenated to an embedding $\textbf{e}_{in}$, conditioning our dynamic NeRF that models the 4D face dynamics of a subject.
% After extracting audio features $\textbf{e}_a$, and expression features $\textbf{e}_e$, we concatenate them to form a new feature $\textbf{e}_{in}$. The feature $\textbf{e}_{in}$ conditions our dynamic NeRF, in order to model the 4D face dynamics of a subject.
%% Moving to suppl
% Similar to AD-NeRF~\cite{guo2021adnerf}, we parse the head from the background using an automatic parsing method, namely MaskGAN~\cite{maskgan}. Further, we assume that the last point of each ray takes the RGB color of the background and lies on it.
% The talking head is represented by an implicit function $F_\Theta$ that corresponds to an MLP. 
For each video frame, we fit a 3DMM~\cite{baselfacemodel, facewarehouse} and extract the head pose and camera parameters, in order to estimate the viewing direction $\textbf{d}$.
% After converting the head pose from the observation space to the canonical space, we use the estimated head pose as the viewing direction $\textbf{d}$ of the radiance field. 
The learned feature $\textbf{e}_{in}$, the viewing direction $\textbf{d}$ and a 3D point location $\textbf{x}$ in canonical space are input to the implicit function $F_{\Theta}$ (MLP), which predicts the corresponding RGB color \textbf{c} and density $\sigma$:
% of the point: 
\begin{equation}
    F_{\Theta}: (\textbf{e}_{in}, \textbf{d}, \textbf{x}) \longrightarrow (\textbf{c}, \sigma)
\end{equation}    
Given the color $\textbf{c}$ and density $\sigma$ at each sampled point of every ray, we can reconstruct each video frame using volumetric rendering. For each camera ray $\textbf{r}(t)=\textbf{o}+t\textbf{d}$, where $\textbf{o}$ is the camera center,
% and $\textbf{d}$ is the viewing direction,
the color $C$ is estimated by accumulating the RGB colors and densities of the points sampled along the ray: $C(\textbf{r};\Theta) = \int^{t_f}_{t_n} \sigma_{\Theta}(\textbf{r}(t)) \textbf{c}_\Theta(\textbf{r}(t),\textbf{d}) T(t) dt$~\cite{mildenhall2020nerf},
% \begin{equation} 
%     C(\textbf{r};\Theta) = \int^{t_f}_{t_n} \sigma_{\Theta}(\textbf{r}(t)) \textbf{c}_\Theta(\textbf{r}(t),\textbf{d}) T(t) dt
% \end{equation}
where $T(s) = \exp(-\int^{t}_{t_n}\sigma_{\Theta}(\textbf{r}(s)))ds$ is the accumulated transmittance from $t_n$ to $t$, and $t_f$ and $t_n$ are the far and near bounds respectively. We denote the outputs of $F_\Theta$ as $\textbf{c}_{\Theta}$ and $\sigma_{\Theta}$ for brevity. 
Similar to~\cite{mildenhall2020nerf}, we learn a coarse and a fine model for hierarchical volumetric rendering. We optimize our NeRF using a photo-consistency loss:
\begin{equation}
    \mathcal{L}_{photo} = \sum_{\textbf{r}\in \mathcal{R}}||\hat{C}(\textbf{r};\Theta)-C(\textbf{r};\Theta)||^2_2 ,
\end{equation}
which measures the mean squared error between the ground truth color $C(\textbf{r};\Theta)$ and the predicted color $\hat{C}(\textbf{r};\Theta)$, and $\mathcal{R}$ is the set of all the rays in each batch (see also suppl.). 
% ~for more details.

% \paragraph{Implementation details} For our complete technique, we used identities from MEAD to train the NeRF. For each identity, we used the highest level emotions videos of ``angry'', ``happy'' and ``sad'', and the only level of ``neutral'' to train the NeRF. Since videos in MEAD are only 4-8 seconds long which are impractically small videos for training NeRFs, we concatenate different videos of the same emotion. Further, since not all videos in the same emotion for the same identities were captured in the same exact pose, a requirement for NeRFs to train well, videos of an identity were filtered on the basis of whether they were shot immediately following the other and the focal length was the same and the pose distribution of the faces after fitting a 3DMM model on these videos were very similar. After concatenation, each video was  $\approx 15$ seconds long for each emotion. While training our method, the frames from videos of each emotion were randomly picked with their corresponding audio features $e_a$, expression features $e_e$ (by spanning a window of features of size $\omega$ around the corresponding index) and pose. 

% The model architecture for the dynamic NeRF is the same as AD-NeRF except that the input size changes from $32$ (audio features only) to 96 (audio features + expression features).

% \end{document}

\section{Experiments}
\label{experiments}
% \documentclass[../main.tex]{subfiles}
% \graphicspath{{\subfix{../images/}}}
% \begin{document}

\begin{figure*}[t]
  \begin{center}
  \includegraphics[width=\textwidth]{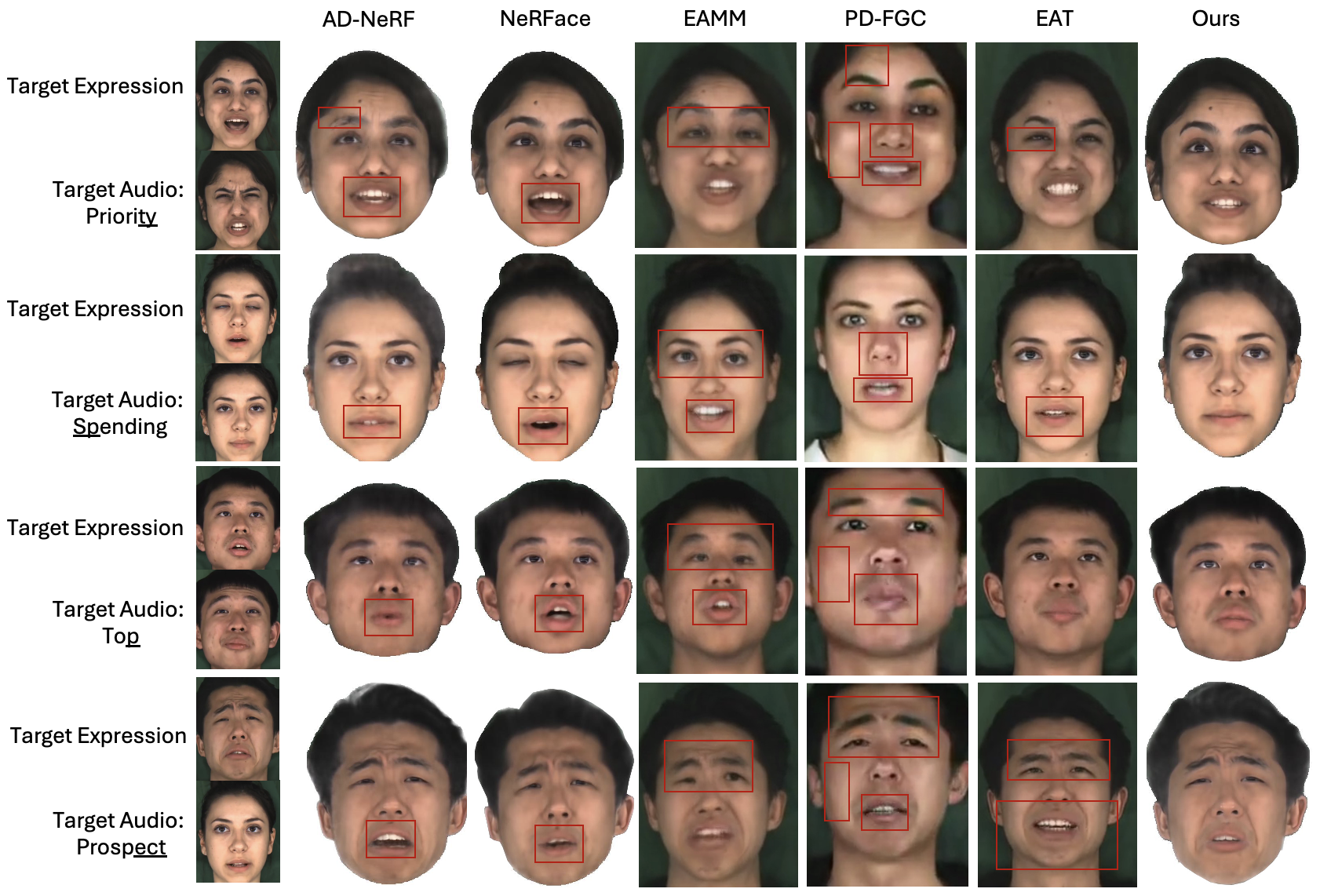}
  \caption{Talking face generation guided by target expression and audio sources (1st column). We compare with state-of-the-art methods for expression and audio-driven talking face generation (EAMM~\cite{jieamm}, PD-FGC~\cite{wang2022pdfgc}), categorical emotion based talking face generation (EAT~\cite{EAT_gen}), as well as the audio-only AD-NeRF~\cite{guo2021adnerf}, and expression-only NeRFace~\cite{nerface}. Our method outperforms all these methods, transferring the expression and audio inputs with higher fidelity, while preserving the target identity.}
  % \caption{Illustration of expression transfer in talking faces as compared to the state of the art. Our method beats the state-of-the-art in terms of expression accuracy and identity consistency, while also improving the lip-sync. AD-NeRF is a strictly audio based model, so no expression source was used for it. NeRFace is a strictly expression based model, so no audio source was used for it. AD-NeRF, NeRFace were trained on the same dataset as ours, as they are identity specific methods. }
  \label{fig:expr-guid}
  \vspace{-18pt}
  \end{center}
\end{figure*}

% \subsection{Dataset}

\textbf{Dataset.}
In our experiments, we use the MEAD dataset~\cite{kaisiyuan2020mead}. MEAD includes 48 identities, performing 7 emotions at 3 intensity levels and 1 neutral emotion. The videos are captured by 7 cameras at different viewpoints. Each emotion level contains $\approx 30-40$ videos corresponding to an utterance sampled from a superset of sentences.  
% for individual identity videos and training our landmark encoder, audio encoder and expression encoder.
% For the self-supervised audio encoder learning, we use the complete set of frontal-view videos of MEAD.
To train our audio encoder (see Sec.~\ref{sec:method_audio}), we use the complete set of frontal-view videos.
% Since MEAD is not consistent in the sentences that are spoken in different emotions we could only use videos with corresponding pairs present in other emotions are used to train the expression disentangler.
For the expression transformer (see Sec.~\ref{sec:method_exp}), we need video pairs, where a person 
% of the same identity
pronounces the same utterance with different emotions. We use the highest level (level 3) of the emotions ``angry'', ``happy'' and ``sad'', and the single level of ``neutral'', leading
% There are 18 sentences among these 4 emotions
% 4 emotions angry (level 3), happy (level 3), sad (level 3) and neutral (level 1), 
% that are uttered in at least 2 emotions for each identity. This leads 
to a total of 84 unique pairs of sentences being spoken in 2 emotions. Of those, 60 pairs of videos are used for training, and 24 for validation. Since each person's expressions
% for a particular emotion while speaking
% are often identity-specific, so we train an expression transformer for each identity
are unique, we train an expression transformer for each identity. 
% Since not all the utterances are spoken in different emotions, we filter MEAD. For each identity, we use 60 pairs of videos for training, and 24 pairs for validation.
% This filtering leads to 30 unique sentences in the resultant dataset. For each identity that was tested from MEAD, 60 pairs of videos were available from the complete MEAD dataset for training. Further, 24 pairs of videos formed the validation set.
% For our complete technique, we used identities from MEAD to train the NeRF. 
To train our dynamic NeRF, we focus on 4 identities from MEAD, training the network for each identity.
% We again use the highest level emotion videos of ``angry'', ``happy'' and ``sad'', and the single level of ``neutral'' for each identity.
Since the videos in MEAD are only 4-8 seconds long, 
% which are impractically small videos for training NeRFs,
we concatenate videos of the same emotion for each identity.
Not all the videos of the same emotion for an identity were captured in the same head pose, so we filter videos based on the estimated focal length and the pose distribution after fitting a 3DMM. After concatenation, we get videos of at least 12 seconds per emotion. 
% For each identity, we train a NeRF by randomly picking frames from all the corresponding emotion videos. 
See suppl. for more details.
% Further, not all videos in the same emotion for the same identities were captured in the same exact pose, a requirement for NeRFs to train well. So, videos of an identity were filtered on the basis of the focal length and the pose distribution of the faces after fitting a 3DMM model on these videos. After concatenation, each video was be at least 12 seconds long for each emotion. While training our method, the frames from videos of each emotion were randomly picked with their corresponding audio features, expression features and pose.

\subsection{Results}
% \begin{table*}[]
% \begin{center}
% \begin{tabular}{lrrrrrr}
%         & \multicolumn{1}{l}{LSE-D↓} & \multicolumn{1}{l}{LSE-C↑} & \multicolumn{1}{l}{Exp-Diff↓} & \multicolumn{1}{l}{PSNR↑} & \multicolumn{1}{l}{SSIM↑} & \multicolumn{1}{l}{LPIPS↓} \\
%         \hline
% AD-NeRF [1] & 10.216                    & 4.380                     & 0.080                        & 20.467                   & 0.698                    & 0.308                     \\
% NeRFace [2] & 12.761                    & 1.918                     & 0.066                        & 19.683                   & 0.672                    & 0.332                     \\
% EAMM [3]   & 12.901                    & 1.771                     & 0.111                        & 18.474                   & 0.592                    & 0.239                     \\
% PDFGC [4]  & 8.445                     & 6.220                     & 0.089                        & 21.094                   & 0.648                    & 0.228                     \\
% EAT [5]    & \textbf{7.736}            & \textbf{7.200}            & 0.079                        & 19.222                   & 0.691                    & 0.334                     \\
% Ours    & 9.899                     & 4.466                     & \textbf{0.043}               & \textbf{21.224}          & \textbf{0.720}           & \textbf{0.207}           
% \end{tabular}
% \end{center}
% \end{table*}
\begin{table*}[t]
\caption{Quantitative comparison of our method with the state-of-the-art. 
% We train a new model following what is proposed in AD-NeRF \cite{guo2021adnerf} and NeRFace \cite{nerface} and use the pretrained models for the rest.
Results are highlighted as follows: \colorbox[HTML]{FF8A33}{Best}, \colorbox[HTML]{FFE993}{Second Best} and \colorbox[HTML]{ADD8E6} {Third Best}.}
\label{tab:comp-table}
\begin{center}
\scalebox{0.9}{
\begin{tabular}{lrrrrrrrrrr}
Method & \multicolumn{1}{l}{LSE-C $\uparrow$} & \multicolumn{1}{l}{ACD $\downarrow$} & \multicolumn{1}{l}{Exp-Diff $\downarrow$} & \multicolumn{1}{l}{PSNR $\uparrow$} & \multicolumn{1}{l}{SSIM $\uparrow$} & \multicolumn{1}{l}{LPIPS $\downarrow$} \\
\hline
AD-NeRF~\cite{guo2021adnerf} & 4.380 & \colorbox[HTML]{ADD8E6}{0.141} & 0.080 & 20.467 & \colorbox[HTML]{ADD8E6}{0.698} & \colorbox[HTML]{ADD8E6}{0.187}\\
NeRFace~\cite{nerface} & 1.966 & \colorbox[HTML]{FFE993}{0.103}& \colorbox[HTML]{FF8A33}{0.023}& \colorbox[HTML]{FF8A33}{21.335}& \colorbox[HTML]{FF8A33}{0.736}& \colorbox[HTML]{FFE993}{0.175}\\
\hline
EAMM~\cite{jieamm} & 1.771 & 0.284 & 0.111 & 18.474 & 0.592 & 0.248 \\
PD-FGC~\cite{wang2022pdfgc} & \colorbox[HTML]{FFE993}{6.220}& 0.657 & 0.089 & \colorbox[HTML]{ADD8E6}{21.094}& 0.648 & 0.228 \\
EAT~\cite{EAT_gen}& \colorbox[HTML]{FF8A33}{7.200}& 0.183 & \colorbox[HTML]{ADD8E6}{0.078}& 19.222 & 0.691 & 0.205 \\
Ours& \colorbox[HTML]{ADD8E6}{4.466}& \colorbox[HTML]{FF8A33}{0.095}& \colorbox[HTML]{FFE993}{0.043}& \colorbox[HTML]{FFE993}{21.224}& \colorbox[HTML]{FFE993}{0.720}& \colorbox[HTML]{FF8A33}{0.174}
\end{tabular}
}
\end{center}
\vspace{-15pt}
\end{table*}

In this section, we compare our proposed method with the state-of-the-art. We include comparisons with AD-NeRF~\cite{guo2021adnerf} that takes only audio as input and NeRFace~\cite{nerface} that takes only 3DMM expression parameters as input, in order to illustrate our method's performance against NeRF-based methods with just one of the inputs, audio or expression. Then, we compare with EAMM~\cite{jieamm}, PD-FGC~\cite{wang2022pdfgc}, and EAT~\cite{EAT_gen} that are identity-generic methods for emotion-aware talking face video generation, trained on large datasets. Note that while EAMM and PD-FGC use a video source for expressions, EAT only takes an emotion label as input to produce expressions. Our method achieves the best disentanglement between expression and audio sources, producing high-quality expressive talking faces. 

\subsubsection{Qualitative Evaluation}
Fig.~\ref{fig:expr-guid} shows our qualitative results. Notice how our method produces accurate lip shapes that follow the target audio (\eg phoneme ``t'' in row 2), while also synthesizing the input expression (\eg sad) with higher fidelity than the other methods.
% We first compare with AD-NeRF, a state-of-the-art NeRF that takes only audio as input, and NeRFace, a state-of-the-art NeRF that takes only expression as input. Even though combining both inputs is more challenging, since it requires appropriate disentanglement, our method achieves better lip-sync accuracy and  expression transfer than those. For example, in the first row, our method produces the lip shape following the corresponding sound ``f'', while also synthesizing the input angry expression with a higher fidelity than the other methods. Next, we compare with EAMM, PD-FGC and EAT, which are the state-of-the-art methods in emotional talking face video generation.
EAMM generally distorts the input face, adds asymmetrical artifacts, and is unable to produce accurate mouth shapes in all rows. While PD-FGC performs better than EAMM in terms of lip-shape accuracy, it still distorts the input identity and produces artifacts. For example, we observe glossy faces, color distortions and lip artifacts in all rows,  
% does a poor job at transferring expressions from the source. For example, in all rows for the PD-FGC column, we see produced expressions depicting similar emotion but a different, more airbrushed, expression as compared to the source. We also see glossy face and lip artifacts in all rows of the PD-FGC column,
a plain white band in place of teeth in rows 1 and 2, and loss of identity-specific characteristics, such as the mole on the face of the woman in row 1. EAT performs best among the other methods, creating accurate lip shapes, synced to the input audio, while also being faithful to the source emotions. However, EAT still struggles with preserving the input identity. For example, it generates artifacts in the eye region and eyebrows in rows 1 and 4 respectively, and wide jaws and crossed eyes in row 4.
% 1, 2, 4, 5 and 6. 
% EAMM fails to make mouth shapes close to the shape required for the sound, PD-FGC and EAT do a good job at producing accurate lip-shapes. 
In general, we observe identity inconsistency problems in all EAMM, PD-FGC, and EAT.
% are not identity-specific, they result in identity inconsistency problems. We suspect that the artifacts we see in the produced images by these methods are from training samples (such as glossy lips in all examples of PD-FGC), as these artifacts are not seen anywhere in the input identity at test time. 
% Moreover, NeRF-based methods, such as AD-NeRF, NeRFace and our method are the only methods that preserve the identity of the person.
In contrast, the NeRF-based methods, \ie AD-NeRF, NeRFace, and our method, learn to preserve the input identity.
% are the only methods that preserve the  identity
% of the person.
Our method demonstrates high-quality results, transferring the source facial expression and following the source audio with higher fidelity. 

\subsubsection{Quantitative Evaluation}

\textbf{Evaluation Metrics.}
We conduct quantitative evaluation on common metrics used in the talking face generation field. We use LSE-C \cite{wav2lip} to measure the lip synchronization of our method, and Peak Signal-to-Noise Ratio (PSNR), Structural Similarity (SSIM)~\cite{ssim} and Learned Perceptual Image Patch Similarity (LPIPS)~\cite{lpips} to measure the image quality against the expression source images. We also estimate the expression transfer accuracy (Exp-Diff)~\cite{wang2022pdfgc} by using 3D face reconstruction~\cite{deep3D_pytorch} and calculating the Mean Squared Error (MSE) of the extracted expression parameters in the synthesized images with those of the driving expression images. Further, we estimate the identity preservation using the Average Content Distance (ACD) metric, inspired by~\cite{sda}, by calculating the cosine distance between ArcFace~\cite{deng2019arcface} face recognition embeddings of synthesized images and driving expression images. Essentially, the idea is that the smaller the distance between those embeddings, the closer are the synthesized images to the driving images in terms of identity.

We show the corresponding quantitative results in Table \ref{tab:comp-table}.
% We compare our method against AD-NeRF and NeRFace which are audio only and 3DMM expression parameters only methods respectively. 
Since Exp-Diff and the visual quality metrics are computed against expression source frames, we find that the expression-only NeRFace performs best on those metrics. Our method significantly outperforms the state-of-the-art in emotion-aware talking face generation (EAMM, PD-FGC, EAT) in terms of visual quality (PSNR, SSIM, LPIPS), identity preservation (ACD), and expression transfer (Exp-Diff).
% , such as PSNR, SSIM and LPIPS. We also find that our method outperforms the state-of-the-art in terms of preserving the identity of the input, as measured by ACD. Further, we find our method to most faithfully produce expressions from the expression source, as indicated by Exp-Diff. 
% Our method outperforms the state-of-the-art in terms of identity consistency (ACD) and expression difference (Exp-Diff). 
While PD-FGC and EAT do perform better than our method in terms of lip-syncing, as they are trained on large-scale video data, our method outperforms the rest of the methods. We encourage the readers to watch our 
%suppl.~document 
suppl.~video for additional results demonstrating the efficacy of \MethodName.
%, as well as watch our suppl.~video.
% our ablation studies, demonstrating the efficacy of our self-supervised audio encoder training and expression disentanglement, as well as to watch our suppl.~video.
% in terms of lip-sync.
% PD-FGC and EAT's superior performance in terms of lip-sync can be explained by the reasoning that these methods are not identity-specific and thus they use large-scale video data for training.
% to learn accurate associations between lip motion and audio.   
%Whereas, our NeRF based method overfits to the audio that it has seen during training hence making an average face in cases where there are multiple mouth positions that can produce a sound, or behave erratically in out-of-distribution settings. 

\subsubsection{Ablation Study}
\begin{table}[]
\caption{Ablation study on our proposed audio and expression representations. In (a), we train the network without the self-supervised audio encoder learning and without expression transformer (we use features from the pre-trained ResNet-based emotion recognition network from~\cite{EMOCA:CVPR:2021} passed through a thin MLP). In (b), we again omit the self-supervised audio encoder learning and use 3DMM expression parameters. In (c), we add the self-supervised audio encoder learning, but we use expression features as in (a). In (d), we train our expression transformer on 3DMM expression parameters. Best results are highlighted in \textbf{bold}.}
\label{tab:ablation-table}
\begin{center}
\scalebox{0.75}{
\begin{tabular}{lllll|rrr}
\multicolumn{5}{c|}{Method} & \multicolumn{3}{c}{Metrics}                      \\
\hline
\multicolumn{1}{p{1.75cm}}{Variant} & \multicolumn{1}{p{1.75cm}}{Self Supervised Audio Encoder} & \multicolumn{1}{p{1.75cm}}{3DMM Expression Parameters} & \multicolumn{1}{p{1.75cm}}{Emotion Recognition Features} & \multicolumn{1}{p{1.75cm}|}{Expression Transformer} & \multicolumn{1}{l}{LSE-C$\uparrow$} & \multicolumn{1}{l}{Exp-Diff$\downarrow$} & \multicolumn{1}{l}{LPIPS$\downarrow$} \\
\hline
(a)& &      & $\checkmark$    &      & 1.804          & 0.027          & 0.170 \\
      % \hline
(b) & & $\checkmark$    &      &      & 1.848          & \textbf{0.022} & \textbf{0.161} \\
      % \hline
(c) & $\checkmark$     &      & $\checkmark$    &      & 1.760          & 0.028          & 0.168          \\
% \hline
(d) & $\checkmark$     & $\checkmark$    &      & $\checkmark$    & 2.982          & 0.071          & 0.190          \\
% \hline
Full Net. & $\checkmark$     &      & $\checkmark$    & $\checkmark$    & \textbf{4.466} & 0.043          & 0.174         
\end{tabular}
}
\end{center}
\vspace{-5pt}
\end{table}
% We conduct ablation studies on our method as follows. 

% \subsubsection{Expression Disentanglement} 

\textbf{Expression Disentanglement.} 
% We analyze the impact of expression-disentanglement here.
Table \ref{tab:ablation-table} shows different variants of our method, demonstrating the efficacy of our self-supervised learning of our audio representation, as well as the disentanglement of our expression representation. More specifically, in variant (a), we omit the self-supervised audio encoder learning and the expression transformer (we directly use features from a pre-trained ResNet-based emotion recognition network from \cite{EMOCA:CVPR:2021} mapped through a thin MLP, and the audio encoder is trained along with the NeRF). In variant (b), we use 3DMM \cite{baselfacemodel} expression parameters and skip the self-supervised audio encoder. In variant (c), we add the self-supervised audio encoder learning, but we use expression features as in (a). Finally, in variant (d), we learn the expression features, using 3DMM
expression parameters as input to our transformer. We see that not disentangling the emotion recognition features in (a) and (c), and the expression parameters from~\cite{baselfacemodel, facewarehouse} in (b), cause the NeRF network to only learn expressions from the expression source and ignore the audio source. This leads to a significant decrease in lip-sync metrics on unseen audio and best performance in terms of Exp-Diff and LPIPS. Further, trying to disentangle 3DMM expression parameters in (d) fails to learn meaningful features which leads to poor lip-sync metrics on unseen audio and the worst performance in terms of Exp-Diff and LPIPS. Our proposed expression transformer leads to a successful disentanglement between expression and speech-specific lip motion.
% The inclusion of expression-disentanglement enables self-supervised audio encoder learning to work. 
% The NeRF completely overfits to 3DMM expression parameters from, hence resulting in the best \textit{Exp-Diff}. 
Note that Exp-Diff is computed against the driving expression images, which implies that if the network has overfitted to the driving expression
% generated images are only overfit to the driving expression,
the corresponding Exp-Diff would also be lower. 
Disentangling expressions from lip motion leads to a balanced performance between expression and lip-sync accuracy.

\noindent
\textbf{Interpreting Learnt Features.}
% We conduct further analysis to investigate the proposed expression disentanglement and the nature of the learned expression features.
% To illustrate the necessity of using the expression disentanglement we conduct further analysis shown in Fig.~\ref{fig:interp} and Fig.~\ref{fig:tsne-plot}.
\begin{figure}
  \centering
  \includegraphics[width=\linewidth]{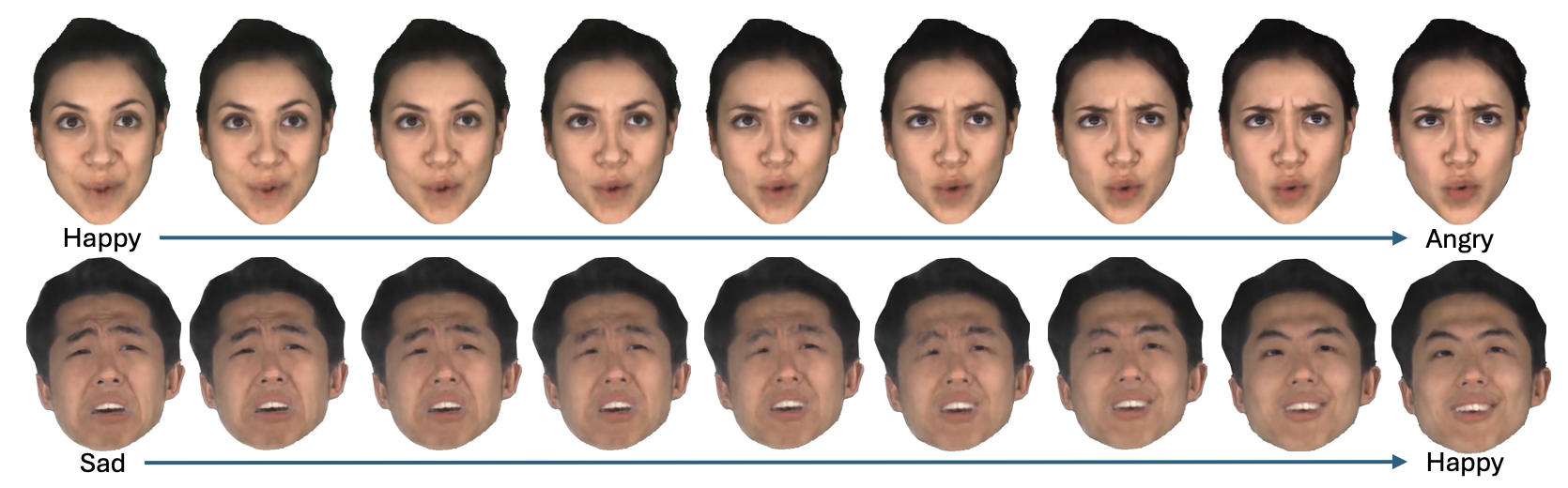}
  \caption{Additional analysis that shows that the expression encoder disentangles features that are semantically grounded and well-behaved. Interpolation of features between different emotional expressions leads to semantically meaningful expressions.}
  \label{fig:interp_main}
  \vspace{-10pt}
\end{figure}
In Fig.~\ref{fig:interp_main}, we conduct further analysis to investigate the proposed expression disentanglement and the nature of the learned expression features. The interpolation result between two expression features, learned by our expression transformer, shows that our method learns semantically grounded features.
In the suppl.~material, we show additional interpolation results and show t-SNE plots of the learned expression features, indicating that they are semantically meaningful. 
\vspace{-10pt}

\section{Conclusion}
\label{discussion}
% \documentclass[../main.tex]{subfiles}
% \graphicspath{{\subfix{../images/}}}
% \begin{document}

% We show expression-guided NeRF-based talking face generation. We argue that previous methods for expression transfer in talking-face generation have lack in realism and non-trivially change the appearance of the person while achieving good lip-sync. Our method achieves better expression transfer than all other methods while also improving on the lip-synchronization as compared to NeRF based state of the art methods. Moreover, quantitative and qualitative experiments demonstrate that our method can synthesize high-fidelity talking head videos.

In conclusion, we introduce a novel method for joint expression and audio-guided talking face generation. Prior work either struggles to preserve the speaker identity or fails to synthesize faithful facial expressions. We propose a self-supervised method to extract audio features, aligned to lip motion, achieving accurate lip synchronization to unseen audio. In addition, we design a transformer-based module to learn expression features, disentangled from speech-specific mouth motion. By conditioning on the learned representations, our dynamic NeRF synthesizes high-fidelity talking face videos, providing simultaneous control of facial expressions and lip movements, and
% , synthesizing high-fidelity talking face videos.
% that follow the respective audio and expression sources, 
outperforming the current state-of-the-art.
We argue that our proposed representations can be easily extended to other neural rendering pipelines, such as Gaussian Splatting~\cite{kerbl3Dgaussians}, that we plan to explore as future work.

\noindent
% \blfootnote{\textbf{Acknowledgements.} This work was partially supported by Amazon Prime Video.}
\textbf{Acknowledgements.} This work was supported in part by Amazon Prime Video and a grant from the CDC/NIOSH (U01 OH012476).

\bibliography{egbib}
\clearpage
\section*{Supplementary}

\subsection*{Contents}

\noindent
The supplementary is organized as follows: 
\begin{enumerate}
    \item Additional Ablation Study in Sec. \ref{sec:ablation}
    \item Additional Results in Sec. \ref{sec:addnal}
    \item Implementation Details in Sec. \ref{sec:impl_det}
    % \item Details on Results in Sec. \ref{sec:test_det} 
    \item Discussion: Limitations and Ethical Considerations in Sec. \ref{sec:discussion}
\end{enumerate}
We strongly encourage the readers to watch our supplementary video.

\section{Additional Ablation Study}
\label{sec:ablation}

\begin{figure}
    \centering
    \begin{subfigure}[b]{0.69\textwidth}
        \centering
        \includegraphics[width=\textwidth]{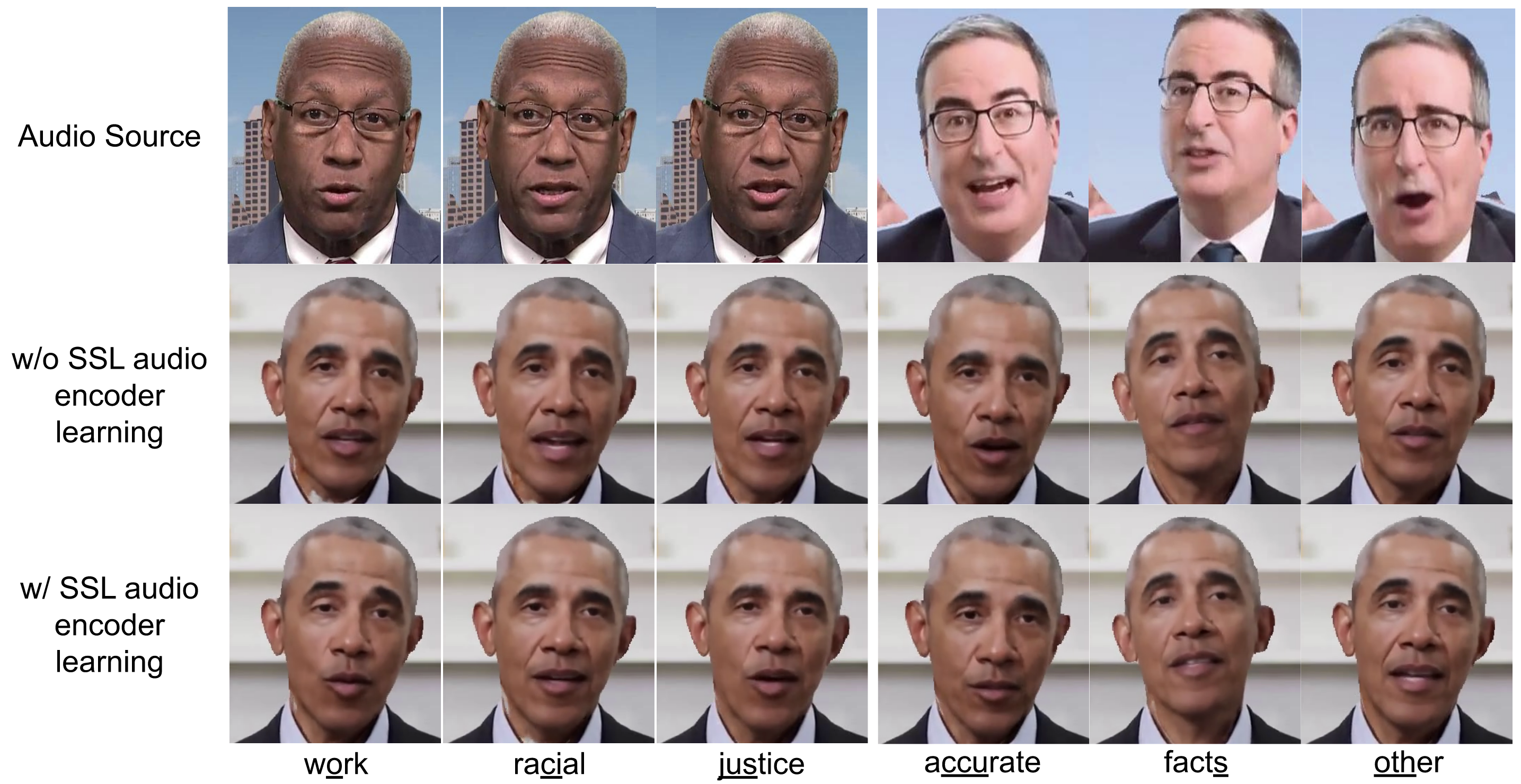}
        \caption{}
    \end{subfigure}
    \hfill
    \begin{subfigure}[b]{0.29\textwidth}
        \centering 
        \includegraphics[width=\textwidth]{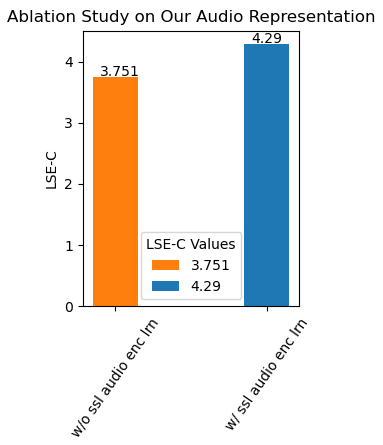}
        \caption{}
    \end{subfigure}
\caption{(a) illustrates the ablation with and without self-supervised audio encoder learning. Our audio encoder learning produces better and more accurate mouth shapes as compared to without it. In (b) we quantitatively evaluate our method without and with self-supervised audio encoder learning.}
\label{fig:lip-sync}
% \vspace{-16pt}
\end{figure}

\noindent
\textbf{Audio Encoder Learning.} We conduct an ablation study on the impact of the self-supervised audio encoder learning. We find that our proposed audio representation improves the lip-sync performance. We show this impact in Figure~\ref{fig:lip-sync}(b), where the method with self-supervised audio encoder learning has better lip-sync performance than the method without it. We conduct this experiment without including expression input. Figure~\ref{fig:lip-sync}(a) illustrates the performance of our audio encoder learning qualitatively. Our proposed self-supervised learning leads to more accurate mouth shapes, corresponding to the spoken phonemes. Note that we omit the expression disentanglement during this evaluation and evaluate it on Obama2 and John Oliver from the datasets collected in LSP~\cite{obama2_lsp} and the PATS dataset~\cite{pats-1, pats-2}.
% word being spoken.
% , than without it. 
% Figure \ref{fig:lip-sync} shows that our method makes more realistic mouth shapes than other methods.

\begin{figure}
  \centering
  \includegraphics[width=\linewidth]{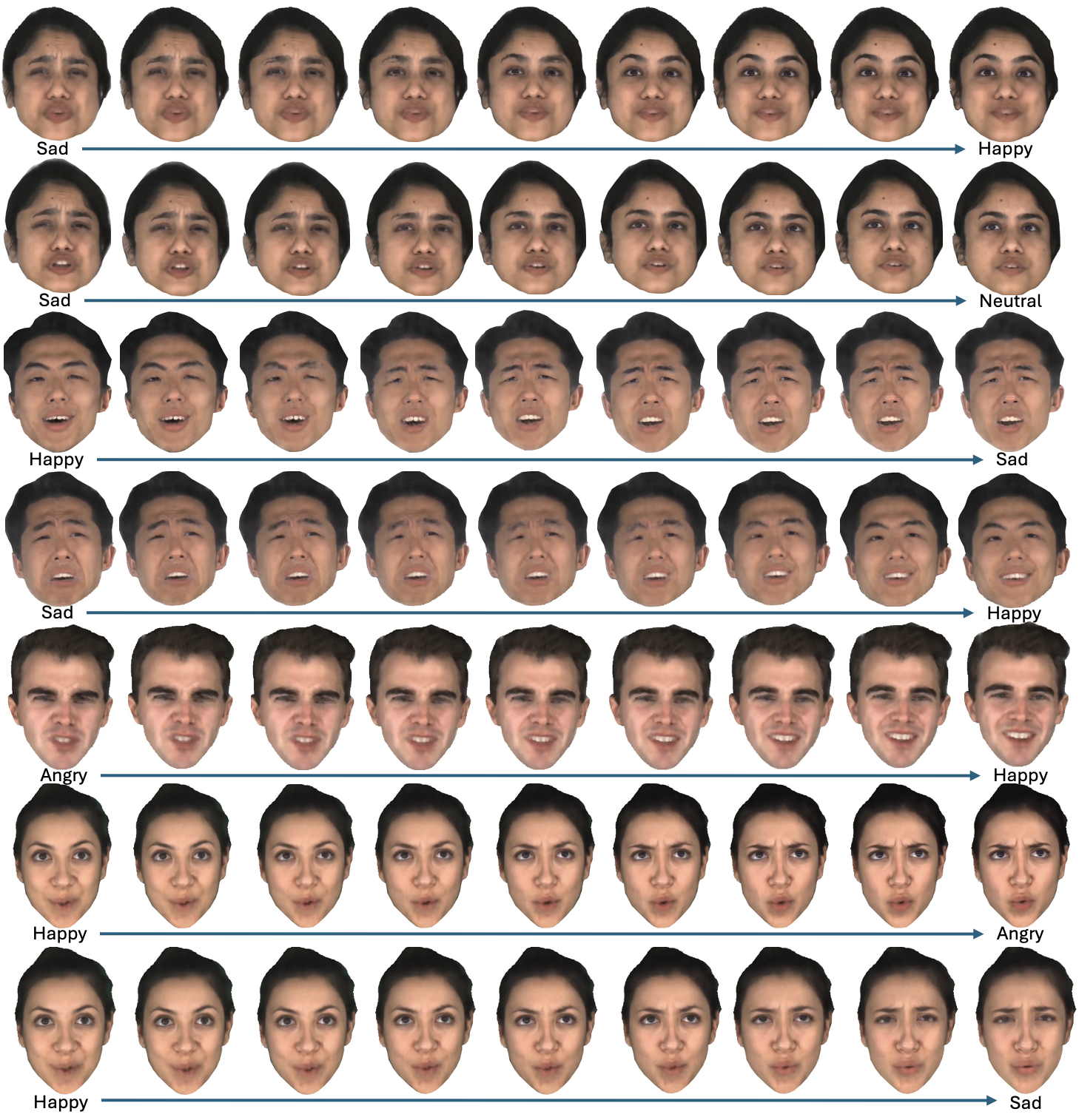}
  \caption{Additional analysis that shows that the expression encoder disentangles features that are semantically grounded and well-behaved. Interpolation of features from left to right leads to semantically meaningful expressions.}
  \label{fig:interp}
  
\end{figure}

\begin{figure}
  \centering
   \includegraphics[width=\linewidth]{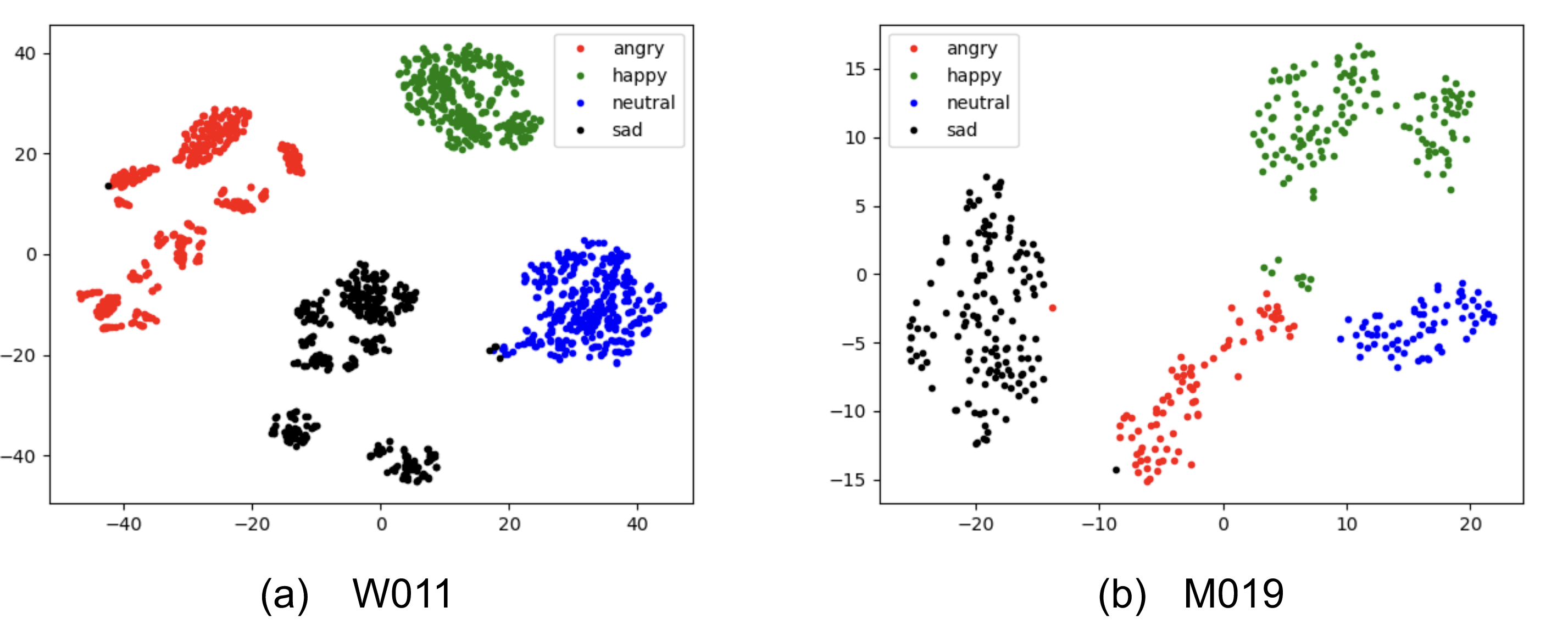}
  \caption{t-SNE plots of the expression features learned by our expression transformer for the ``W011'' and ``M019'' identities of MEAD.}
  \label{fig:tsne-plot}
\end{figure}

\noindent
\textbf{Interpreting Learnt Features.}
% We conduct further analysis to investigate the proposed expression disentanglement and the nature of the learned expression features.
% To illustrate the necessity of using the expression disentanglement we conduct further analysis shown in Fig.~\ref{fig:interp} and Fig.~\ref{fig:tsne-plot}.
In Figure~\ref{fig:interp}, we provide additional examples of interpolation between two expression features, learned by our expression transformer and representing two different emotions. 
% These are generated by giving corresponding interpolation results to the trained NeRF, using the same input audio features. 
As shown, 
% from our expression transformer and inputs these features to the trained NeRF while keeping the audio features the same. It shows that 
our method learns semantically grounded features. In Figure~\ref{fig:tsne-plot}, we demonstrate t-SNE plots of the expression features learned by our proposed transformer for two identities. The plots indicate that the learned features are well-behaved and clustered together according to the corresponding emotion annotations.
% learned by the expression encoder are well-behaved and cluster together. 

% \begin{Figure}
%   \centering
%   \includegraphics[width=\linewidth]{images/Interp.png}
%   \caption{Additional analysis that shows that the expression encoder disentangles features that are semantically grounded and well-behaved. Interpolation of features from left to right leads to semantically meaningful expressions. }
%   \label{fig:interp}
% \end{Figure}

\section{Additional Results}
\label{sec:addnal}
\begin{figure*}[t]
  \includegraphics[width=\textwidth]{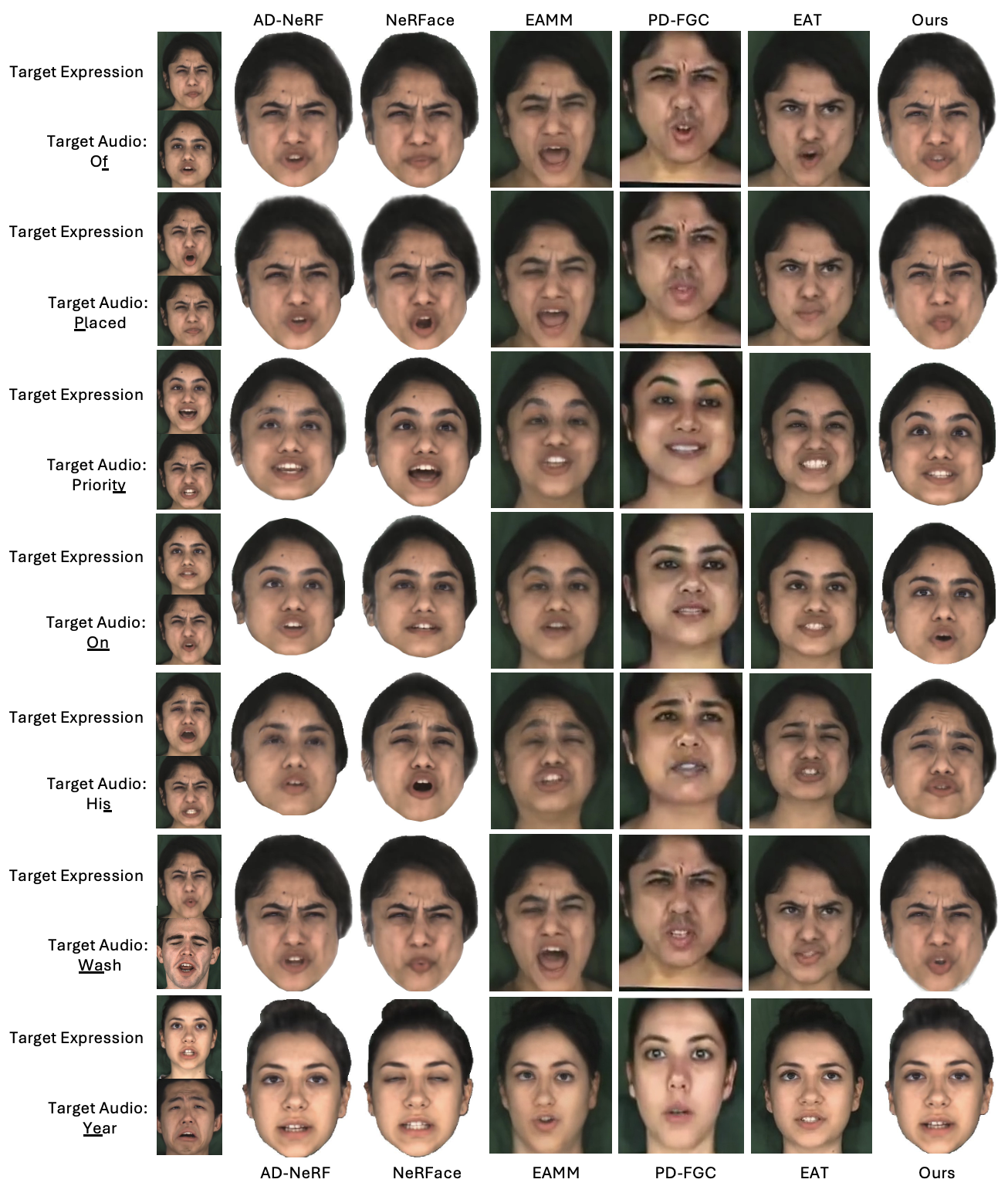}
  \caption{Talking face generation guided by target expression and audio (1st column) compared with the state-of-the-art.}
  \label{fig:expr-guid-1}
\end{figure*}

\begin{figure*}[t]
  \includegraphics[width=\textwidth]{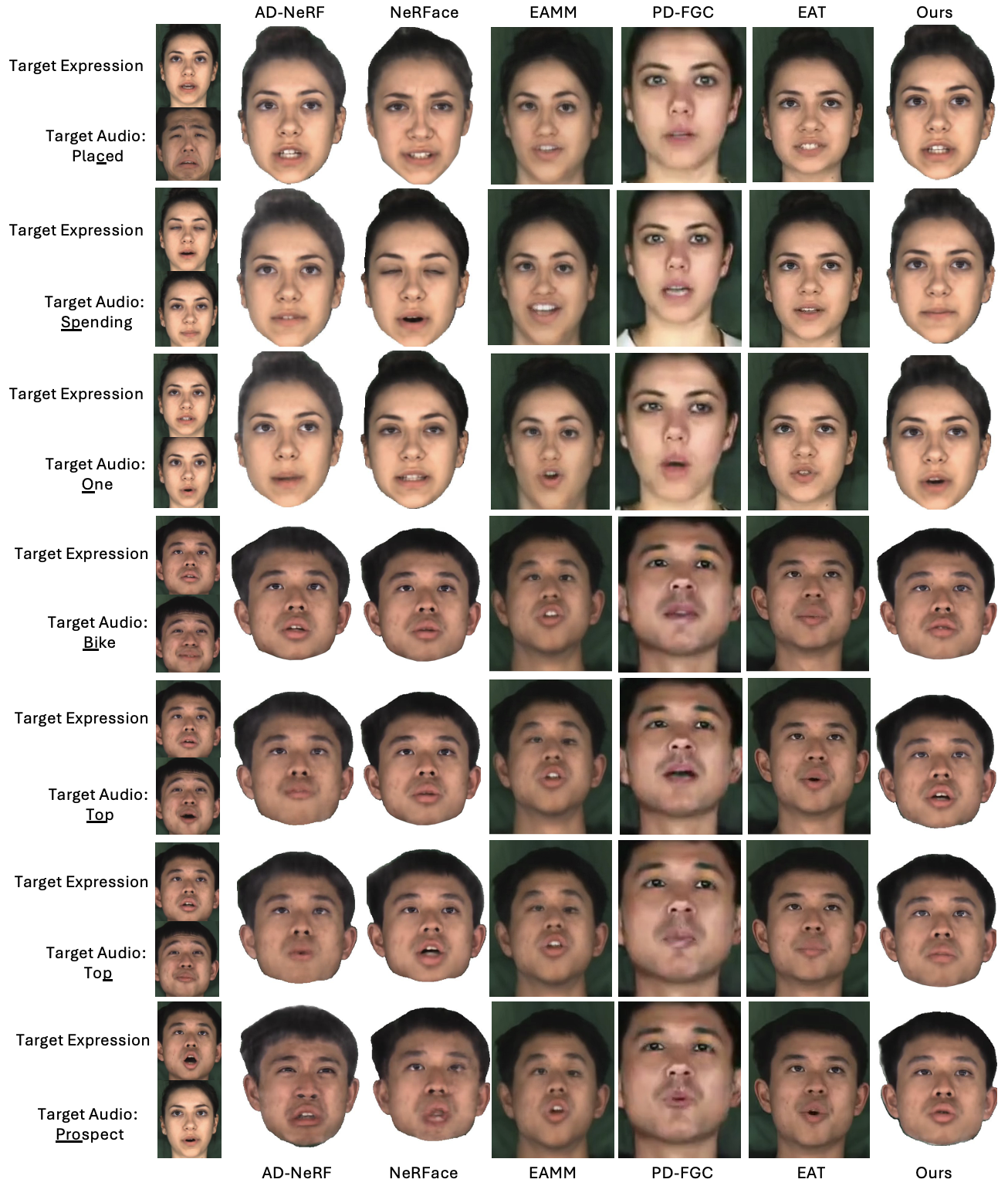}
  \caption{Talking face generation guided by target expression and audio (1st column) compared with the state-of-the-art.}
  \label{fig:expr-guid-2}
\end{figure*}

\begin{figure*}[t]
  \includegraphics[width=\textwidth]{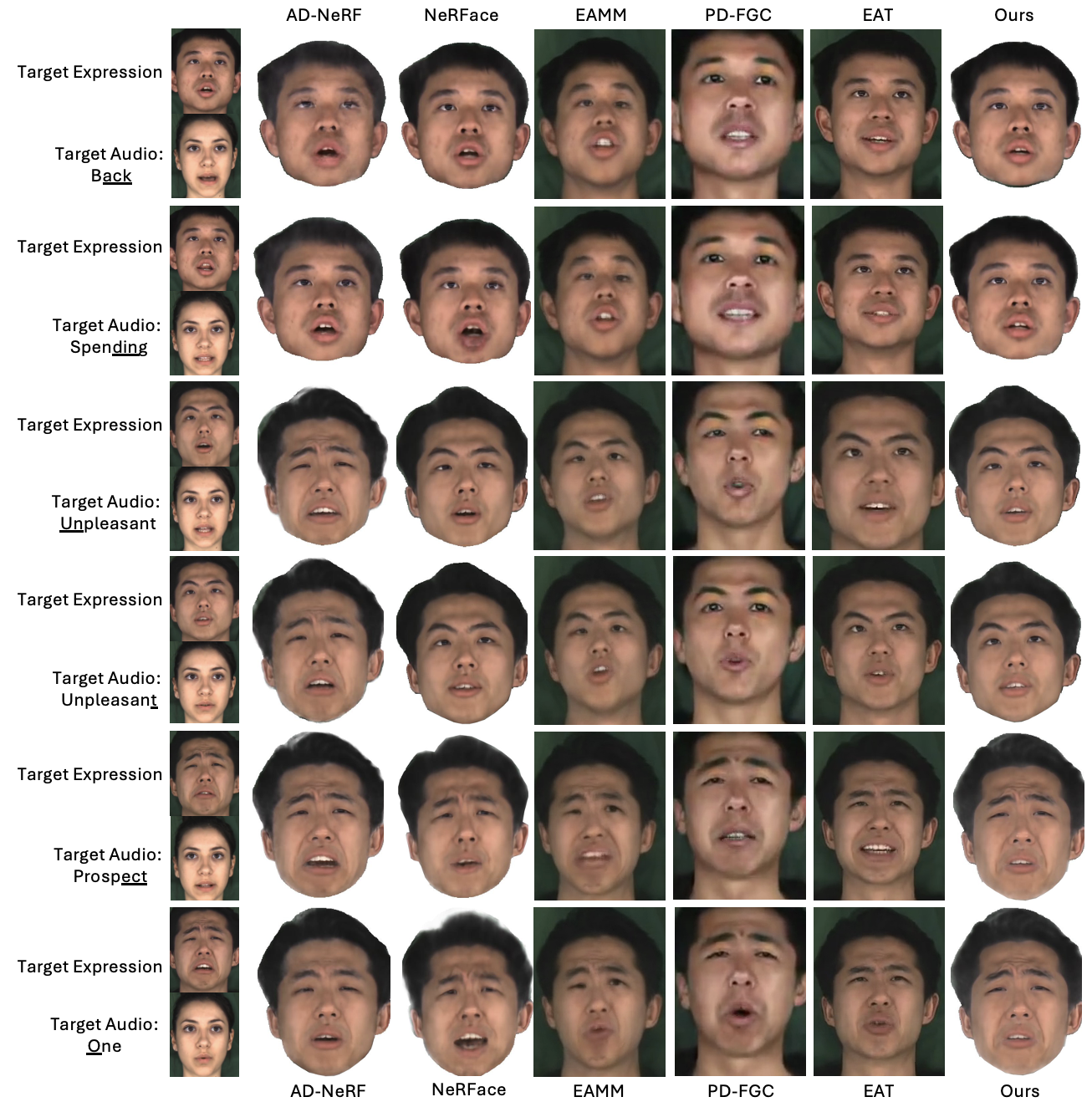}
  \caption{Talking face generation guided by target expression and audio (1st column) compared with the state-of-the-art.}
  \label{fig:expr-guid-3}
\end{figure*}

In Figures~\ref{fig:expr-guid-1}, ~\ref{fig:expr-guid-2}, ~\ref{fig:expr-guid-3} we show additional results of rendered frames in comparison with state-of-the-art methods for expression and audio-driven talking face generation (EAMM~\cite{jieamm} and PD-FGC~\cite{wang2022pdfgc}), categorical emotion based talking face generation (EAT~\cite{EAT_gen}) as well as the audio-only AD-NeRF~\cite{guo2021adnerf}, and expression-only NeRFace~\cite{nerface}. We also compare with  Our method outperforms all the other methods, transferring the expression and audio inputs with a high fidelity, while preserving the target identity.

\section{Implementation Details}
\label{sec:impl_det}
\subsection{Landmark Autoencoder}
\subsubsection{Dataset}
For this task, we use the complete available set of frontal view MEAD~\cite{kaisiyuan2020mead} dataset videos, \textit{i.e.} ``part0''. $39$ identities from the dataset were kept for training and rest $9$ for validation. We used 68 point landmarks of which 17 landmarks corresponding to the face contour were discarded for training the landmark autoencoder and these were input to the autoencoder as arrays. We extract landmarks from square frame of $480\times480$ rescaled and cropped from the initial resolution of $1920\times1080$ (by cropping the middle $1080\times1080$ frame from the video and then resizing it to $480\times480$).
% a square of edge size $720\times720$.
% During training, the frames were are randomly sampled from the same video of an identity to keep the feature disentanglement identity and video emotion agnostic. 
\subsubsection{Pipeline}
The landmark encoder and decoder consisted of 4 MLPs each with 256, 256, 128 as the hidden layer sizes and Leaky ReLU activations (negative slope set to 0.02). The output features $e_f$ and $e_m$ were 64 dimensions each. For swapping, $\epsilon$ was set to $0.8$ and the last 20 points in the landmarks representing the lips were used.

 We noticed that sampling the frames $A$ and $B$ from different videos of different identities or emotions would encourage the network to learn features that may contain identity and emotion information. Essentially, the network would try to differentiate between identity-specific or emotion-specific characteristics of the face, rather than learn lip motion features. This would deteriorate the performance of our dynamic NeRF, which requires conditioning on a powerful audio representation. Thus, we sample frames $A$ and $B$ randomly from the same video of an identity.
% We noticed that sampling from different videos of different identities or emotion led to the network learning features that may contain identity and emotion information. This was because the network would have to learn such information in order to handle swapped reconstruction. We found that using such features to condition the audio encoder would lead to deteriorated performance in the final NeRF output. So during training, the frames $A$ and $B$ are randomly sampled from the same video of an identity.

\subsubsection{Training}
 The model was optimized using Adam~\cite{kingma2014adam} with default parameters and $10^{-5}$ as weight decay value. We trained the landmark encoder for 15 epochs on a single Nvidia V100 GPU with a batch size of 256 which takes $\approx12$ hours.

\subsection{Contrastive Learning}

\subsubsection{Dataset}
For this task, the complete available set of frontal view videos from MEAD dataset was used, \ie from ``part0''. We use the same splits used for training the landmark autoencoder, \ie $39$ identities for training and rest $9$ for validation. We extract landmarks from square frame of $480\times480$ rescaled and cropped from the original resolution of $1920\times1080$ (by cropping the middle $1080\times1080$ frame from the video and then resizing it to $480\times480$). The corresponding audio signals for each video was converted into DeepSpeech~\cite{deepspeech} features.

\subsubsection{Pipeline}
We use the trained encoder from our Landmark Autoencoder and keep its weights frozen during training. We use the same architecture as the audio encoder of \cite{guo2021adnerf} for our audio encoder. During training, negative audio samples were randomly selected from the same identity but different video. Further, the temperature $\tau$ in the InfoNCE loss formulation in Eq. 2 was set to $0.1$.

\subsubsection{Training} The audio encoder was optimized using Adam with learning rate set to $10^{-3}$ and default parameters for the rest with an exponential decay and final learning rate to be $5 \times 10^{-6}$. We trained this model for 30 epochs on a single Nvidia V100 GPU with a batch size of 256 which takes $\approx2$ hours.

\subsection{Expression Transformer}
\subsubsection{Dataset}
We use the MEAD dataset as it contains identities speaking the same utterances in different emotions. This dataset is not consistent in the sentences that are spoken in different emotions and hence only videos of an utterance with corresponding videos of the same utterance present in other emotions are used to train the expression transformer. This filtering leads to $18$ unique sentences for each identity. For each identity from MEAD that was used to test in our method, $60$ pairs of videos where each video in the pair speak the same sentence with different emotions were used for training. The rest ($15-24$) of the pairs of videos formed the validation set. Note that we only included the highest level videos of the emotions ``angry'', ``happy'' and ``sad'', and the only level of ``neutral'' for these pairs. 

For each video in these pairs, we extract the emotion features $e_{emo}$ per video frame, using a vanilla ResNet50-based emotion recognition network from~\cite{EMOCA:CVPR:2021} trained on AffectNet~\cite{affectnet} on expression classification, valence, and arousal regression task jointly. We use the features from the ResNet backbone before being processed by any classification/regression heads. To account for any misalignment of the utterances of different emotions, we incorporate the dynamic time-warping (DTW) algorithm~\cite{1104847} to align the input emotion feature sequences.

\subsubsection{Pipeline}
We find that each person's style of speaking with a particular emotion is unique, so training a separate expression encoder is necessary. 
% This pretraining of the expression encoder is identity specific, as each person is unique in their ways of speaking with a particular emotion. 
For the expression autoencoder, we used a transformer architecture with linear layers in the beginning to map down the $2048$ dimensional emotion recognition features to $128$ via two MLP layers with Leaky ReLU activation and a hidden layer size of $512$. The output of the encoder is split in half resulting in $64$ dimension features for lip motion and expressions. The encoder had $8$ attention heads, $3$ layers with dropout enabled and set to $0.2$. A positional encoding was applied similar to \cite{transformer} and since a decoder was used, the input sequence was padded with start and end sequence vectors. The expression decoder had the same architecture in terms of the number of attention heads and layers. The start and end sequence tokens were stripped off and the features were mapped back to $2048$ dimensions using a 2 layer MLP with Leaky ReLU activations (negative slope set to 0.02) and hidden layer size of $512$. $\omega$ is set to $8$. Further, $\delta$ is set to $0.8$. During training, all features output from the encoder corresponding to each input feature are passed through to the decoder after the swapping operation. 
% However, during inference, we only use the middle feature (5th feature in window of 8 features) to represent the window. 
\subsubsection{Training}
The expression transformer was optimized using Adam with default parameters and weight decay of $10^{-5}$. We trained this model for 20 epochs on a single Nvidia V100 GPU with a batch size of 256 which takes $\approx1$ hour.  

\subsection{NeRF} 

\subsubsection{Dataset}
We used identities from MEAD to train the NeRF. For each identity, we used the highest level emotions videos of ``angry'', ``happy'' and ``sad'', and the only level of ``neutral'' to train the NeRF. Since videos in MEAD are only 4-8 seconds long which are impractically small videos for training NeRFs, we concatenate different videos of the same emotion. Further, all videos in the same emotion of one identity were not captured in the same pose or at the same distance from the camera.
To ensure that the NeRF is not overfitted to specific emotion - head pose or emotion - camera distance pairs, we filter the videos of each identity.
% Using such videos to train our NeRF would cause blurry and shift artifacts. 
% This is because the NeRF would struggle to render the face at a particular distance from the camera and with a certain pose reliably as it has seen these factors vary in training.  
% a requirement for NeRFs to train well
More specifically, videos of an identity were filtered on the basis of three factors: (1) whether they were shot immediately following each other (determined by the lack of sudden change in pose and distance from camera if two videos were concatenated), (2) whether the determined focal length for the face after fitting a 3DMM~\cite{baselfacemodel, facewarehouse} was the same, and (3) whether the pose distribution were very similar. After concatenation, each video of a particular emotion was  $\approx 15$ seconds long.

\subsubsection{Pipeline}\label{sec:implementation_pipeline_3dmmfitting}
The talking head is represented by an implicit function $F_\Theta$ that corresponds to an MLP. The model architecture for this implicit function is the same as AD-NeRF, consisting of 8 linear layers with a hidden size of 128 and ReLU  activations. However, the input size was changed from $32$ (audio features only) to 96 (audio features + expression features). 
For each video frame, we fit a 3DMM~\cite{baselfacemodel, facewarehouse} and extract the head pose and camera parameters, in order to estimate the viewing direction $\textbf{d}$. After converting the head pose from the observation space to the canonical space, we use the estimated head pose as the viewing direction $\textbf{d}$ of the radiance field.
As in AD-NeRF, we positionally encode 3D point \textbf{x} with 10 frequencies and viewing direction \textbf{d} with 4 frequencies but not features $e_e$ and $e_a$.
While training the NeRF, the frames from videos of each emotion were randomly picked with their corresponding audio features $e_a$, expression features $e_e$ (by spanning a window of features of size $\omega$ around the corresponding index and selecting the $\omega/2$th feature of the output as $e_e$) and pose. 
Similar to AD-NeRF~\cite{guo2021adnerf}, we parse the head from the background using an automatic parsing method, namely MaskGAN~\cite{maskgan}. Further, we assume that the last point of each ray takes the RGB color of the background and lies on it. During inference, we render the talking face on a white background.
% we replace the background with a white image and render the talking face on it. 

\subsubsection{Training}
The NeRF was optimized using Adam with an initial learning rate $5\times10^{-4}$ that decays exponentially to $5\times10^{-5}$. At each iteration, we 
randomly sample 2048 rays for a video frame. The model is trained for 400,000 iterations (around 2 days on a single GPU). We also enable the fine-tuning of the audio encoder during this training with a learning rate of $5\times10^{-6}$. 

\subsection{Metrics}
We compute the visual quality metrics, namely PSNR, SSIM~\cite{ssim}, LPIPS~\cite{lpips}, after extracting a face bounding box using 3DDFA V2 ~\cite{3ddfa_cleardusk, guo2020towards} and cropping the face. The target frame from the expression source is also cropped in the same manner and resized to the size of the cropped source frame. This is done so that the background is not included in the quality computation. The identity preservation metric (ACD) is measured using ArcFace~\cite{deng2019arcface}, a ResNet50-based network trained on WebFace~\cite{zhu2021webface260m}. Specifically, we used ``buffalo\_l'' model from the insightface repository~\cite{insightface}. For lip-sync confidence~\cite{wav2lip} of NeRFace, we compute the metric of each video against all testing audio sources and report the average. Similarly, for Exp-Diff~\cite{wang2022pdfgc} for AD-NeRF, we compute the metric of a particular video against all testing expression source videos and report an average.

% \section{Details on Results}
% \label{sec:test_det}
% \input{supplementary_sections/testing}

\section{Discussion}
\label{sec:discussion}
\subsection{Limitations}

An important factor of our method is the 3DMM fitting that is used as ground truth to extract the head pose and camera parameters (see Sec.~3.3 of the main paper and Sec.~\ref{sec:implementation_pipeline_3dmmfitting}). This fitting can be noisy and the error can be propagated to the final generated videos. Improving the face tracking further would be an interesting future work. In addition, we propose a NeRF-based representation that synthesizes high-quality expressive talking faces. We believe that our proposed audio and expression representations can be easily extended to other neural rendering pipelines, like 3D Gaussian Splatting~\cite{kerbl3Dgaussians} that provides faster training and inference times. We plan to explore this direction in the future.

% Our NeRF based talking-face generation method performs worse than adversarial \cite{jieamm} \cite{wang2022pdfgc} methods in lip-synchronization because NeRFs overfit on the train data. Due to this overfit, unseen audio often causes weird behaviour because it is out-of-distribution for the model. This causes two problems: 1. an average face when multiple face shapes can explain the same sound; 2. artifacts when the sound is out-of distribution. These can be solved by taking the lip-synchronization feature space closer to the visual domain (eg. landmarks) and using large amounts of data (eg. VoxCeleb2). While our method works for certain identities, it also struggles to disentangle expressions from lip-motions properly due to improper alignment by the dynamic time-warping technique. This is illustrated by fig. \ref{fig:M003-tsne} which shows poor clusters. This can be improved by using phoneme based alignment. 
% Sometimes we notice that our NeRF might overfit to the audio that it has seen during training, hence making an average face in cases where there are multiple mouth positions that produce a sound.
% \begin{figure}[]
%   \centering
%   \includegraphics[width=0.5\linewidth]{M003_transformer_emotion_tsne.png}
%   \caption{t-SNE plots of the expression features using our expression encoder for ``M003'' identity in MEAD.}
%   \label{fig:M003-tsne}
% \end{figure}

\subsection{Ethical Considerations}

We would like to note the potential misuse of generation methods. With the rise of ``deep fakes'', it becomes easier to make photorealistic fake videos of any speaker. These can be used for malicious purposes, e.g. to spread misinformation. To this end, it is important to develop accurate methods for fake content detection and forensics~\cite{cai2022marlin, reiss2023detecting}. We intend to share our source code to help improving such research. In addition, appropriate procedures must be followed to ensure fair and safe use of videos if used for training or inference. In our work, we use MEAD that is a publicly available dataset, featuring actors talking with different emotions in a controlled setup and has been used by several other published works.

\end{document}